\documentclass[lettersize,journal]{IEEEtran}
\usepackage{amsmath,amsfonts}
\usepackage{algorithmic}
\usepackage{algorithm}
\usepackage{array}
\usepackage[caption=false,font=normalsize,labelfont=sf,textfont=sf]{subfig}
\usepackage{textcomp}
\usepackage{stfloats}
\usepackage{url}
\usepackage{verbatim}
\usepackage{graphicx}
\usepackage{cite}
\usepackage{bbm}
\usepackage{hhline}
\usepackage{boldline}
\usepackage{multirow}
\usepackage{pifont}
\usepackage{xcolor}
\usepackage{booktabs}
\usepackage{makecell}
\usepackage{wrapfig}
\usepackage{orcidlink}

\usepackage{xspace}
\makeatletter
\DeclareRobustCommand\onedot{\futurelet\@let@token\@onedot}
\def\@onedot{\ifx\@let@token.\else.\null\fi\xspace}

\newcommand{\revtwo}[1]{\textcolor{black}{#1}}

\newcommand{\revone}[1]{\textcolor{black}{#1}}

\begin{document}

\title{Improving Visual Object Tracking 
\\ through Visual Prompting}


\author{Shih-Fang Chen\orcidlink{0000-0002-8438-400X}, 
Jun-Cheng Chen\orcidlink{0000-0002-0209-8932}~\IEEEmembership{Member,~IEEE}, 
I-Hong Jhuo\orcidlink{0009-0009-3893-3758}~\IEEEmembership{Member,~IEEE},
\\and Yen-Yu Lin\orcidlink{0000-0002-7183-6070}~\IEEEmembership{Senior Member,~IEEE}

\vspace{-0.18in}
\thanks{
Manuscript received March 23, 2024; revised May 27, 2024 and August 12, 2024; accepted September 20, 2024. Date of publication: month day, 2024;.
This work was supported in part by the National Science and Technology Council (NSTC) under grants 112-2221-E-A49-090-MY3, 111-2628-E-A49-025-MY3, 112-2222-E-001-001-MY2, and 112-2634-F-002-006-, and by Academia Sinica under grant AS-CDA-110-M09. We also thank to National Center for High-performance Computing (NCHC) of National Applied Research
Laboratories (NARLabs) in Taiwan for providing computational and storage
resources. (Corresponding author: Jun-Cheng Chen.)}
\thanks{S.-F. Chen and Y.-Y. Lin are with the Department of Computer Science, National Yang Ming Chiao Tung University. J.-C. Chen is with the Research Center for Information Technology Innovation, Academia Sinica, Taiwan. I-H. Jhuo is with Microsoft, Seattle, Washington, United States.
}
%

}

\markboth{IEEE TRANSACTIONS ON MULTIMEDIA}%
{Shell \MakeLowercase{\textit{et al.}}: A Sample Article Using IEEEtran.cls for IEEE Journals}
%
%
%
%
\maketitle


\begin{abstract}

Learning a discriminative model that distinguishes the specified target from surrounding distractors across frames is essential for generic object tracking (GOT). Dynamic adaptation of target representation against distractors remains challenging because prevailing trackers exhibit limited discriminative capability.
To address this issue, we present a new visual prompting mechanism for generic object tracking, termed PiVOT. PiVOT introduces mechanisms that leverage the pretrained foundation model (CLIP) to automatically generate and refine visual prompts online, thereby enabling the tracker to suppress distractors through contrastive guidance.
To transfer contrastive knowledge from the foundation model to the tracker, PiVOT automatically propagates this knowledge online and dynamically generates and updates visual prompts. Specifically, it proposes a prompt initialization mechanism that produces an initial visual prompt highlighting potential target locations. The foundation model is then used to refine the prompt based on appearance similarities between candidate objects and reference templates across potential targets. After refinement, the visual prompt better highlights potential target locations and reduces irrelevant prompt information.
With the proposed prompting mechanism, the tracker can generate instance-aware feature maps guided by the visual prompts, which are incrementally and automatically updated during tracking, thereby effectively suppressing distractors.
Extensive experiments across multiple benchmarks indicate that PiVOT, with the proposed prompting mechanism, can suppress distracting objects and improve tracking performance.
%
\revtwo{Code 
is publicly available\footnote{https://github.com/chenshihfang/GOT}.}

\vspace{0.01 in}

\end{abstract}
    
\vspace{-0.1in}

\section{Introduction}
\label{sec:intro}

\IEEEPARstart{G}{eneric} object tracking (GOT) estimates the target object's state in each frame of a streaming video, given its initial state in the first frame.
Learning a discriminative representation of the target object is essential to alleviate interference from distracting objects.
Despite substantial progress in trackers such as DiMP~\cite{DiMP} and SiamRPN++~\cite{SiamRPN++}, representation learning and adaptation remain highly challenging in GOT because only limited target information is available during testing to handle unfavorable variations, such as illumination changes, appearance changes, and occlusions.

The strong generalization requirement of GOT motivates us to investigate whether foundation models, such as CLIP~\cite{CLIP}, which is contrastively trained on 400 million image--text pairs, can benefit tracking.
In particular, we investigate whether the category-level contrastive knowledge provided by a foundation model can be transferred to the instance-aware setting of GOT.
Our method leverages CLIP's strong zero-shot capabilities to compare arbitrary objects for automatic visual prompt refinement.
Accordingly, the proposed approach is designed to handle both seen and unseen objects and to allow the tracker to adapt to new targets.
Although CLIP encodes category-level knowledge, a tracker relies on instance-aware features to distinguish the tracked target from surrounding objects, including other instances of the same category and objects with similar appearance.
To bridge this gap, inspired by recent prompting mechanisms such as SAM~\cite{SAM} and SEEM~\cite{SEEM}, we introduce a new prompting mechanism that makes the tracker promptable through dynamically refined visual prompts.

\begin{figure}[!t] \centering
\includegraphics[width=0.42\textwidth]{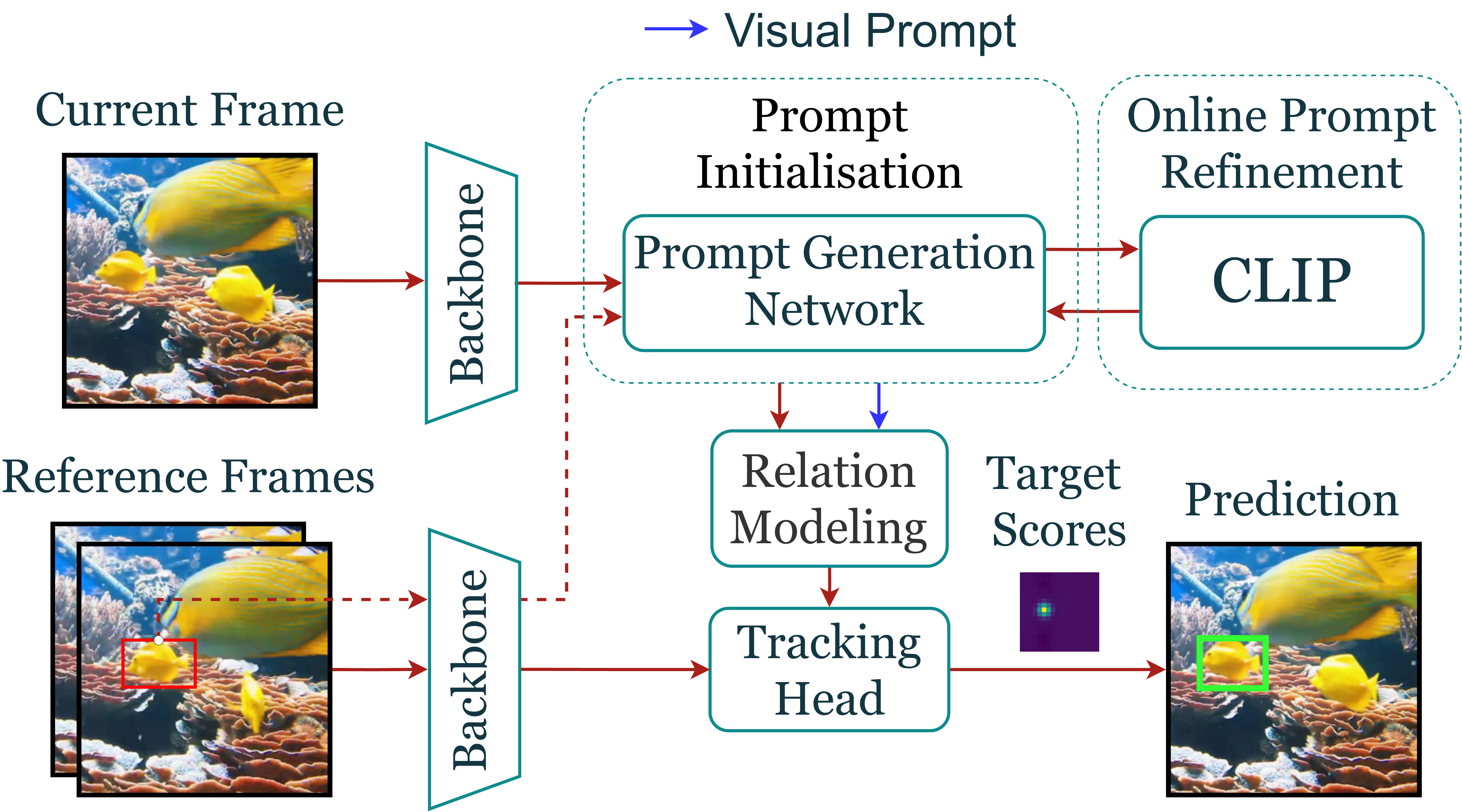}
\vspace{-0.05in}
\caption{
Given the features of the current frame and the reference frames, they are fed into the \emph{Prompt Generation Network} (PGN).
The PGN collaborates with CLIP to enable automatic prompt generation and refinement.
It generates an initial visual prompt that highlights target candidates.
The strong zero-shot recognition capability of CLIP for arbitrary objects enables effective discrimination between the target and distractors among these candidates.
This capability is further exploited to refine the visual prompt.
The \emph{Relation Modeling} module processes the features of the current frame together with the visual prompt to generate enhanced features for the current frame.
The Tracking Head then processes the refined current-frame features and the reference-frame features to produce the tracking prediction.
}
\label{fig:teaser2}
\end{figure}

Figure~\ref{fig:teaser2} illustrates the proposed tracker, which takes the current frame and the reference images as input.
After feature extraction by the backbone network, the \emph{Prompt Generation Network} (PGN) generates a score map by correlating the feature map of the current frame with those of the templates, namely, the images inside the red boxes in the reference frames.
This score map highlights potential target objects in the current frame and serves as the initial visual prompt in our method.
To enable automatic visual prompt generation and refinement via the PGN with CLIP, we crop several RoIs from the current frame, each corresponding to a potential target object.
CLIP is then used for feature extraction and similarity analysis, based on which the similarities between these RoIs and the templates are evaluated.
RoIs with higher similarity to the templates are emphasized on the score map at their corresponding locations.
The resulting refined score map is referred to as the visual prompt in this work.
Prompt refinement is performed only during inference to improve the tracking of arbitrary objects while reducing training cost.

To enable the tracker to be guided by visual prompts, we propose the \emph{Relation Modeling} module, which takes the visual prompt and the feature map of the current frame as input and suppresses distracting objects by reducing their feature responses.
Guided by CLIP-refined visual prompts, PiVOT effectively distinguishes the target from surrounding objects.
As noted in CAML~\cite{CAML}, images with similar visual and semantic characteristics yield similar CLIP embeddings.
Therefore, the CLIP image encoder enables comparisons among arbitrary, class-agnostic objects.
Consequently, the refined visual prompt inherits category-discriminative capability and improves robustness to appearance and viewpoint changes, thereby enhancing tracking performance.
It also improves robustness to temporary occlusion and erroneous updates by suppressing irrelevant object classes, supporting more stable, continuous tracking.
The Tracking Head then uses the prompted features as part of its input to complete the tracking process.
Overall, the proposed paradigm resembles human visual perception, dynamically performing contrastive analysis between the tracked object and surrounding distractors to enable effective tracking.

Fully fine-tuning a large pretrained model is computationally expensive and can lead to overfitting to labeled data~\cite{SPTNet}.
We therefore extend our method by using a frozen ViT-L backbone with DINOv2~\cite{DINOv2} for feature extraction.
Unlike previous works~\cite{SeqTrack,TATrack,CTTrack,GRM,UVLTrack}, which fine-tune the ViT-L backbone on tracking datasets, our method leverages the attributes of foundation models~\cite{FoundationModels}.
This design allows us to combine the frozen ViT-L backbone with a lightweight adapter, requiring less than 1\% of trainable parameters for feature-extractor adaptation, instead of fine-tuning the full pretrained backbone.
As a result, the proposed method remains training-efficient and further improves tracking performance by leveraging the dense, generalized features produced by the foundation model.

The main contributions are summarized as follows. First, we introduce an automatic visual prompt generation and refinement mechanism that does not require prompt annotation from humans, thereby enabling automatic knowledge transfer from the foundation model through visual prompts.
Second, we propose a prompting mechanism for generic object tracking that enables the tracker to generate feature maps that suppress distractors under visual prompts, thereby improving tracking performance.
Extensive evaluations across multiple tracking benchmarks show that the proposed method, aided by CLIP-based visual prompts, effectively improves the tracker's discriminative capability.
As a result, PiVOT substantially improves tracking performance over baseline methods.


\section{Related Work}
\label{sec:related}

\textbf{Generic Object Tracking.}
Generic 
object tracking (GOT) aims at estimating the state of an arbitrary target in a video sequence, given its initial state in the first frame.
Building a robust GOT model is a significant challenge despite extensive research. 
The literature on GOT~\cite{javed2022visual} is extensive.
We focus on relevant paradigms, such as Siamese network-based trackers, DCF-based trackers, and their emerging transformer variants.


Siamese trackers
such as
SiamRPN~\cite{SiamRPN} and SiamRPN++~\cite{SiamRPN++}, 
take a target template and a search image, compute features, and use cross-correlation to create a response map. 
These trackers learn features offline on a large volume of data without online adaptation during inference. 
Despite their effectiveness, they struggle to generalize to new targets and search images dissimilar to training data.


DCF trackers, 
particularly those based on model prediction,
such as 
DiMP~\cite{DiMP} and ToMP~\cite{ToMP},
use the paired support and query images
to generate discriminative correlation filters online through meta-learning
~\cite{
Hypernetworks, 
SNAIL
}.
%
The resultant filters can better identify the target from the background, resulting in higher generalization capabilities of these trackers.
For instance, DiMP adopts this DCF paradigm and outperforms most Siamese trackers. 
Based on model prediction-based DCF trackers,
our tracker further leverages a foundation model by the proposed prompting mechanism to explore the target areas, which helps derive more discriminative filters. 


Tracking models like STARK~\cite{STARK}, TransT~\cite{TransT}, and TrDiMP~\cite{TrDiMP} employ the attention mechanism in Transformers to construct Tracking Heads and showcase superior performance.
The follow-up research efforts leverage Transformers for Tracking Head construction and image feature extraction.
%
Mixformer~\cite{MixFormer} employs a Vision Transformer (ViT) equipped with a Mixed Attention Module 
as its backbone for feature extraction, while SeqTrack~\cite{SeqTrack} employs a masked autoencoder (MAE) pre-trained ViT-L backbone.
They cast regression as a sequence prediction task, predicting boxes sequentially and autoregressively.
These methods that use Transformers for feature extraction have high computation costs during training since they need to fine-tune heavy Transformers.
%

Our method builds on ToMP~\cite{ToMP} because its paradigm dynamically predicts a model that can prevent overfitting to training data more effectively than other paradigms.
ToMP leverages ResNet and incorporates a Transformers-based model predictor for Tracking Head construction. Unlike ToMP, we develop a visual prompting mechanism where CLIP is leveraged to compute visual features for arbitrary tracking objects,
While the model predictor is adopted to make those objects with indistinguishable appearance instance-aware.
\revone{
Our method can derive more adaptive capability and discriminative features for 
Generic Object Tracking task
based on the proposed prompting mechanism with CLIP-refined visual prompts.}

Unlike prior works~\cite{SeqTrack,DropMAE,UVLTrack,RTS} that fine-tune their backbones with tracking datasets or with both tracking data and additional data such as object segmentation~\cite{Youtube-vos,DAVIS}, action recognition~\cite{K700}, etc., our approach leverages the foundation model characteristics of DINOv2~\cite{FoundationModels}. We construct a feature extractor by integrating the frozen backbone with a lightweight adapter, using merely 1M parameters compared to fine-tuning the 300M-parameter ViT-L backbone for tracker training. In contrast to the previous work~\cite{MAT}, which suggests that freezing the pre-trained backbone during tracker training can hinder performance, we find that training a tracker with a frozen pre-trained foundation model DINOv2 backbone can still enhance tracker performance.


\begin{figure*}[!t]
	\centering

        \includegraphics[width=0.98\linewidth, 
        trim={0cm 0 18cm 0}, clip]{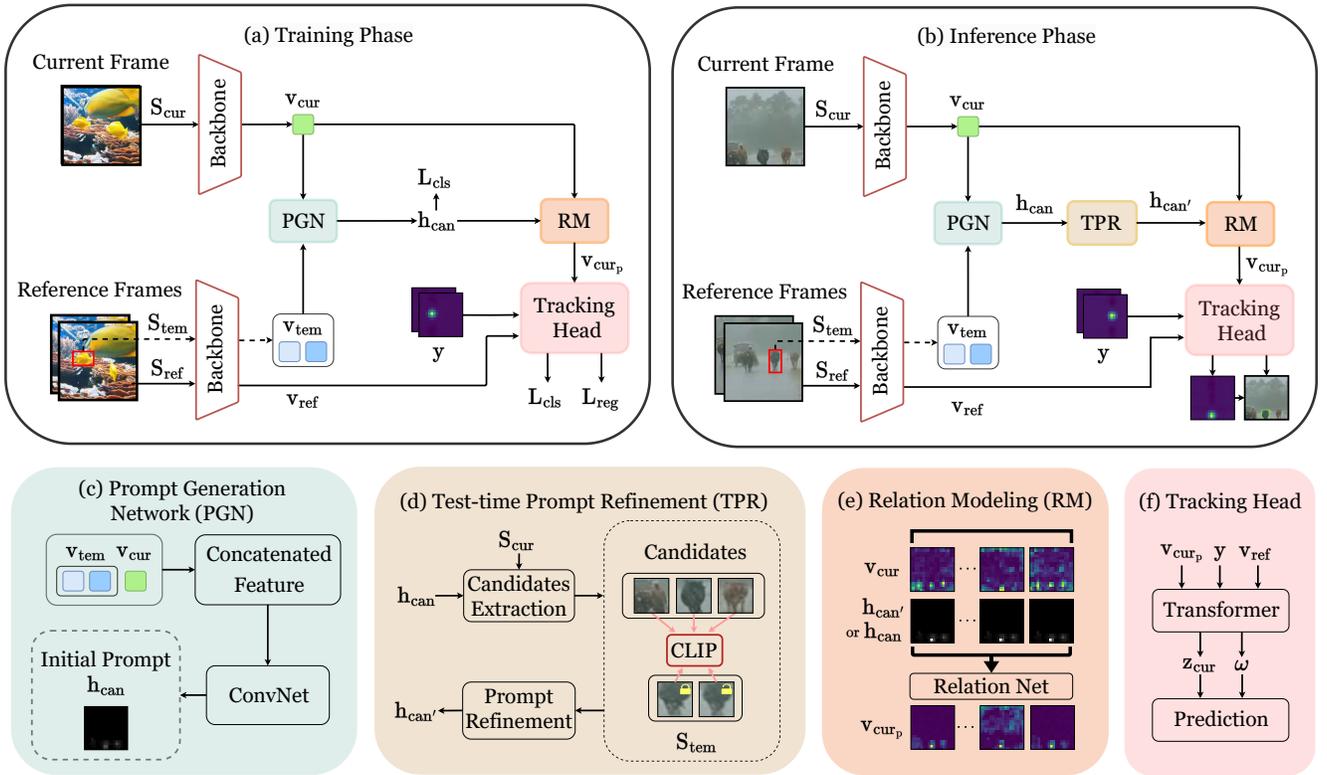}
    
	\vspace{-1.5 cm}
         
	\caption{
        \textbf{ Overview of PiVOT.}
        During the (a) training phase, 
        we aim to make the tracker promptable 
        by introducing (c) 
        \emph{Prompt Generation Network}  (\textbf{PGN})  and 
        (e) \emph{Relation Modeling}  (\textbf{RM})  module. 
        The PGN learns to generate an initial prompt
        %
        \revone{
        and RM enables the tracker to be prompted through the visual prompt.}
        \revone{
        (f) Tracking Head 
        predicts the resultant target state and coordinates.
        }
        During the (b) inference phase, 
         (d) \emph{Test-time Prompt Refinement} (\textbf{TPR}), 
        leverages CLIP to improve the visual prompt, 
        %
        \revone{
        as the zero-shot contrastive ability of CLIP enables it to handle arbitrary tracking objects. 
        Through our proposed components, the visual prompt can be automatically generated and improved via CLIP without the need for human annotation throughout the sequence.
        }
        %
        In the case shown in the figure of RM, 
        a prompt that highlights the target cattle location suppresses distractors through the RM module.
        }
	\label{fig:main2}
\vspace{-0.2 cm}
\end{figure*}


\textbf{Prompting for Tracking.}
Recent research such as OVTrack~\cite{OVTrack}, CiteTracker~\cite{CiteTracker}, 
OneTracker~\cite{OneTracker}, 
and ViPT~\cite{ViPT} introduce the concept of prompt for tracking.
OVTrack, tailored for multi-object tracking (MOT), utilizes knowledge distilled from CLIP and uses a text prompt to enhance tracking. 
While MOT handles objects of predefined classes, GOT typically focuses on an object of an arbitrary class, even unseen during training.
The process of knowledge distillation makes OVTrack concentrate on familiar class-specific features, undermining the broad generalization demands in 
GOT.
\vspace{-0.05 in}

Contemporary vision-language tracking, exemplified by~\cite{divert, CiteTracker,OneTracker}, relies on predefined language descriptors for tracking. 
However, as indicated in \cite{DINOv2}, captions may not sufficiently capture intricate pixel-level details within images.
In contrast, our proposed prompting method exclusively utilizes CLIP image features because, as described in CAML~\cite{CAML}, images with similar visual characteristics and semantic meanings result in similar CLIP embeddings.
%
Additionally, CLIP is not limited to classifying fixed categories but can distinguish between any arbitrary categories.

Recent research, such as SAM~\cite{SAM} and SEEM~\cite{SEEM}, pioneers promptable segmentation. 
%
These methods generate segmentation masks based on various prompts, such as points or boxes. 
Investigations into SAM for tracking lead to SAM-PT~\cite{SAM_PT}, 
emphasizing point tracking~\cite{omnimotion} in video segmentation.
In SAM-PT, SAM is applied during testing with point prompts sourced from prior segmentation masks of SAM. 
Yet, it faces challenges such as appearance and viewpoint variations, as well as background clutter issues, which are critical for GOT tasks. Additionally, it requires a segmentation mask annotation in the initial frame, a feature often absent in most generic object-tracking datasets.
In contrast, in our method, the point prompts are refined by the foundation model CLIP, alleviating 
these issues
by the enhanced discriminative ability.

For prompting interaction and tracking with the foundation model, the 
Generic Object Tracking
task trained with 20 million images, as suggested by ViPT~\cite{ViPT} and OneTracker~\cite{OneTracker}, can also serve as a foundation model tuning task.
While ViPT and OneTracker handle a tracking task demanding depth, thermal infrared, text, and event information prompts, our method uses only RGB data and is inspired by SAM and SEEM.
We introduce a mechanism to pinpoint targets using a visual prompt, which can be automatically refined through CLIP online.
%
While ViPT and OneTracker treat the pre-trained tracker as a foundation model, we further introduce foundation models CLIP and DINOv2, designing mechanisms to seamlessly integrate them into the 
GOT
task to aid tracking.

\vspace{-0.2 cm}

\section{Method}
\label{sec:method}






%
An overview of our method is shown in Figure~\ref{fig:main2}. 
Given the current frame and several reference frames, a backbone network is used for feature extraction. 
The Tracking Head is used to identify the target position in the current frame. 
To make the tracker promptable, we introduce a 
\emph {Prompt Generation Network} (PGN) and a 
\emph {Relation Modeling} (RM) module before the Tracking Head.
The PGN is a weak version of the Tracking Head to generate a score map where the potential target locations in the current frame are highlighted. 
The RM utilizes the resultant score map as the visual prompt to refine the feature map, which serves as the input to the Tracking Head to complete tracking. 
This is the procedure adopted during training.
During inference, one additional module, {\em Test-time Prompt Refinement} (TPR), is inserted between PGN and RM.
TPR, shown in the dashed box, leverages CLIP~\cite{CLIP} to compile features.
Hence, the features of the tracked object become more reliable, particularly for objects unseen during training, allowing RM to generate 
a more reliable feature.
\vspace{-0.3 cm}

\subsection{Revisiting DCF Tracking Paradigm}

\label{s3:revisit_DCF}

In this study, we use 
Transformer tracker
ToMP~\cite{ToMP}, as our foundational tracker for its generality and discriminative capability, though our method PiVOT can work with other trackers.
In ToMP, its Tracking Head is a Transformer-based model predictor, 
consisting of a model predictor for predicting convolution filter weights 
and a target model for score map generation.
ToMP maintains two reference frames: $\left\{\mathit{S_{ref_1}}, \;\mathit{S_{ref_2}}\right\}$. 
$\mathit{S_{ref_1}}$ contains the initial tracking template specified by the user and remains unchanged during tracking. 
$\mathit{S_{ref_2}}$ is derived from the result in the previous frame. 
The reference frames are 
cropped larger than the template.
They encompass both the template and its surrounding area in order to establish a filter for better target-background discrimination.
The templates are denoted as 
$\left\{\mathit{S_{tem_1}}, \;\mathit{S_{tem_2}}\right\}$, with one marked within a red box in the reference frame of Figure~\ref{fig:main2}.

Given the reference frames $\left\{\mathit{S_{ref_1}}, \;\mathit{S_{ref_2}}\right\}$ with their labels
$\left\{\mathit{y_1}, \;\mathit{y_2}\right\}$, and
the current frame $\mathit{S_{cur}}$, 
we compute and respectively denote the frame features by
$\left\{
\mathit{v_{ref_1}}, \;
\mathit{v_{ref_2}}, \;
\mathit{v_{cur}} 
\right\}$, each of which is of resolution $\mathbb{R}^{H\times W\times C}$ where $H\times W$ is the spatial resolution and $C$ is the number of channels.
The model predictor of the Tracking Head takes both the frame features and labels as input, and generates the enhanced feature maps for the current frame $\mathit{z_{cur}} \in \mathbb{R}^{H\times W\times C}$ as well as a weight for the filter $\omega \in \mathbb{R}^{1\times C}$.
This filter $\omega$ is derived to discriminate the target from the background, especially distinguishing similar instances and also pinpointing the target location in the current frame. 
Specifically, convolving the current frame features with this filter yields the score map, namely 
\begin{equation}
\mathit{h_{cls}} = \omega * \mathit{z_{cur}}.
\label{eqn:basic_score_map}
\end{equation}
It follows that performing bounding box regressions generates a dense location prediction map $\mathit{d} \in \mathbb{R}^{H\times W\times 4}$ in the {\em ltrb} bounding box representation~\cite{FCOS}. 
The coordinates with the highest confidence in the score map 
is mapped to the regression score map for bounding box prediction.
The filter weights can also be used to predict the regression score map 
since the reference labels contain both classification and regression information.
%
\revone{We built the Tracking Head following the same implementation as ToMP, including its regression map prediction process and other components. Please refer to the ToMP~\cite{ToMP} for the details.}

%
\vspace{-0.1in}

\subsection{Promptable Visual Object Tracking}
\label{s3:promptable}

In the following, we make the tracker promptable by introducing the 
\emph {Prompt Generation Network} (PGN)
and 
\emph {Relation Modeling} (RM)
so that we can leverage the strong zero-shot capability of CLIP to guide and improve the tracker.

\textbf{Prompt Generation Network (PGN):} 
PGN is derived to generate a score map $\mathit{h_{can}} \in \mathbb{R}^{H\times W}$, where the centers of the candidate targets in the current frame are highlighted.
Namely, once $\mathit{h_{can}}$ is resized to the frame resolution; the highlighted locations are expected to be close to the target center.
The relationship between the current frame $\mathit{S_{cur}}$ and the score map $\mathit{h_{can}}$ is shown in Figure~\ref{fig:main2}.
The output score map of PGN is treated as an initial visual prompt for refining the feature map of the current frame before feeding it into the Tracking Head for tracking. 
%
%
With the features of two templates \(\left\{\mathit{v}_{\text{tem}_1}, \; \mathit{v}_{\text{tem}_2}\right\}\) and the current frame \(\mathit{v}_{\text{cur}}\), 
\revone{each of which is of size \(\mathbb{R}^{H \times W \times C}\), }
the score map is computed by feeding the concatenated features \revone{of size \(\mathbb{R}^{H \times W \times 3C}\)} into a convolutional neural network (CNN) \(\phi(\cdot)\),
where
\begin{equation}
\mathit{h_{can}}=\phi (
\left [
\mathit{v_{tem_1}},\;
\mathit{v_{tem_2}},\;
\mathit{v_{cur}}
\right ]
).
\label{eqn:prompt_test_fea}
\end{equation}
The input $\mathit{v_{tem}}$ to PGN is different from the input $\mathit{v_{ref}}$ to the Tracking Head.
$\mathit{v_{tem}}$ encodes the exact bounding box area of the template, specifically the red box $\mathit{S_{tem}}$ in the reference frame depicted in Figure~\ref{fig:main2}. This is because PGN is designed to identify target candidates that match the template in the current frame.
Additionally, $v_{tem_2}$ is updated by comparing the CLIP similarity among the templates. If the similarity between a new template and the initial template surpasses that of the existing template, we update $v_{tem_2}$.
Conversely, $\mathit{v_{ref}}$ encompasses both the template and its surrounding area. This is because the Tracking Head is designed to produce a filter to distinguish the target from the background, 
even if the targets have indistinguishable appearances.
$v_{ref_2}$ is updated based on the confidence score of $\mathit{h_{cls}}$ following 
previous work~\cite{ToMP,DiMP}.
\revone{
Consequently, the network input to the backbone initially comprises five images: one current frame, two reference frames, and two templates for tracking where the updates occur only for the current frame, the second reference frame, and the second template.
}

\textbf{Relation Modeling (RM):} 
Once the score map $h_{can}$ is obtained, it serves as the initial visual prompt and is channel-wise concatenated with the  feature map of the current frame $\mathit{v_{cur}}$ to generate the prompted feature map via
\begin{equation}
\mathit{v_{cur_p}} = 
g_{\phi} (
\left [
\mathit{h_{can}},\;
\mathit{v_{cur}}
\right ]
),
\label{eqn:can_score_map}
\end{equation}
where $g_{\phi }(\cdot)$ is the relation network classifier~\cite{rel_net} consisting of a Conv-BN-GeLU-based architecture.
Then, we feed the prompted feature $\mathit{v_{cur_p}}$ to the
Tracking Head to compute the final target score map $\mathit{h_{cls}}$.

The original relation network is designed for few-shot learning. 
Given a few examples and a query, the classifier learns to analyze the relationship for each query-example pair.
In this work, we adapt the relation network for tracking tasks, making $g_{\phi }$ to learn to distinguish the relationship between the visual prompt and the image features.
It is worth noting that the PGN and RM, being composed of lightweight and simple networks, introduce little to no additional cost for the tracker while maintaining the same complexity.
\vspace{-0.1in}

\subsection{Offline Training}
\label{s3:Training}
Before presenting the details of how to refine the visual prompt $\mathit{h_{can}}$ by CLIP during the test time, we describe our loss function used in the offline training procedure. 
Similar to other recent end-to-end trained discriminative trackers~\cite{DiMP,ToMP}, we sample the current and reference frames from a video sequence to form training sub-sequences. 
The target classification loss is employed from DiMP~\cite{DiMP}, which is a discriminative loss for distinguishing background and foreground regions. The regression loss is a generalized Intersection over Union loss~\cite{GIOU}. 
The total objective function for the proposed method is:
\begin{equation}
\begin{aligned}
{L_{tot}} 
=  \lambda_{cls}  L_{cls}(\hat{h},\;
h_{cls})
+
\lambda_{can}  L_{cls}(\hat{h},\;
h_{can}) \\
+
\lambda_{reg}  L_{reg}(\hat{d},\;
d)
,
\label{eqn:total_cls_loss}
\end{aligned}
\end{equation}
\noindent where $\lambda_{cls}$, $\lambda_{can}$, and $\lambda_{reg}$ weight the corresponding losses.
$\mathit{h_{cls}}$ and $\mathit{d}$ are the predicted classification and bounding box maps while
$\hat{h}$ and $\hat{d}$ are the ground-truth labels.  
$h_{cls}$ and $h_{can}$ share the same 
label $\hat{h}$, which is similar to the ground-truth $\mathit{y}$ (shown in Figure~\ref{fig:main2}) with a Gaussian kernel process to the center of the target. Using this label with the DiMP loss addresses the data imbalance between the target and the background, as stated in DiMP.
\vspace{-0.1in}


\subsection{Test-time Prompt Refinement}
\label{s3:ttpr}

In the following, we show the details of how to leverage CLIP to refine the visual prompt, i.e., the score map 
$\mathit{h_{can}}$, during the test time for tracking improvement. 
Once the score map $h_{can}$ is obtained, we can identify $N$ target candidates where their positions in the score map are denoted by $\mathit{P = \left\{p_{i}\right\}_{i=\text{1}}^{N}}$ and satisfy the following requirements:
\begin{equation}
\mathit{\phi_{max}(h_{can},\;p_{i}) = \mbox{1} \mbox{ and }  h_{can}(p_{i}) \geq \tau }, \mbox{ for } 1 \leq i \leq N,
\label{eqn:can_extract}
\end{equation}
where $\tau$ represents the confidence threshold. $\mathit{\phi_{max}}$ is an indicator function that returns 1 if the score at $p_i$ is a local maximum in $h_{can}$, and 0 otherwise. The local maxima of $h_{can}$ are identified using the max-pooling operation in a $3 \times 3$ local neighborhood with a stride of 3.

In addition, we retrieve the corresponding bounding box for each target candidate from the bounding box regression map from the Tracking Head of the tracker at the last iteration since the scale changes between two frames are typically not significant~\cite{javed2022visual}.
%
We avoid using PGN for target regression since the input $v_{tem}$ provides coarse information. 
It is utilized to predict the positions of multiple candidates.
Utilizing its features to predict the regression map may not produce optimal results. 
Hence, we choose to use the predictions from the Tracking Head instead.
With the bounding boxes, we can crop the $N$ corresponding candidate RoIs $\left\{\mathit{S_{can_i}}\right\}_{i=1}^N$ from the current frame and extract their features using the image encoder of CLIP along with two reference templates $\left\{\mathit{S_{tem_1}},\; \mathit{S_{tem_2}}\right\}$, as illustrated in Figure~\ref{fig:main2}.
Then, we compute an importance score $D_i$ for each target candidate based on the normalized pairwise cosine similarities between the features of reference templates and the target candidates as follows:
\begin{equation}
\mathit{
D_i = \frac{1}{2} 
\sum\limits_{j=1}^{2}
\frac{\exp({
\cos_{sim}(
\mathit{E_{can_i}}, \;
\mathit{E_{tem_j}}
))
}}
{ 
\sum\limits_{k=1}^{N}
\exp(
{
\cos_{sim}(
\mathit{E_{can_k}}, \;
\mathit{E_{tem_j}})
)
}
}
},
\label{eqn:cos_E2}
\end{equation}
where $\cos_{sim}(\cdot,\cdot)$ represents the cosine similarity metric, and $\{E_{can_i} \in \mathbb{R}^{C}\}_{i=1}^{N}$ and $\{E_{tem_i} \in \mathbb{R}^{C}\}_{i=1}^{2}$ indicate the extracted CLIP features for the target candidates and the reference templates, respectively. 
If the importance score is greater than a threshold $\gamma$, we set its corresponding value in $h_{can}$ as 1 for visual prompt refinement. 
This encourages \emph{\revone{R}elation Modeling} to focus on outputting refined current frame features, particularly in areas where the visual prompt highlights, as the training process has taught RM that emphasizing these locations can enhance tracking.
Finally, we get the final prompt $\mathit{h_{can'}}$ to replace $\mathit{h_{can}}$ in Eq.~\ref{eqn:can_score_map}, which can effectively guide RM to output the enhanced features.


\begin{figure*}[!t]
	\centering
 
        \begin{tabular}{ccc}

        \hspace{-0.3cm}
        
        {{\includegraphics[scale=0.22]
        {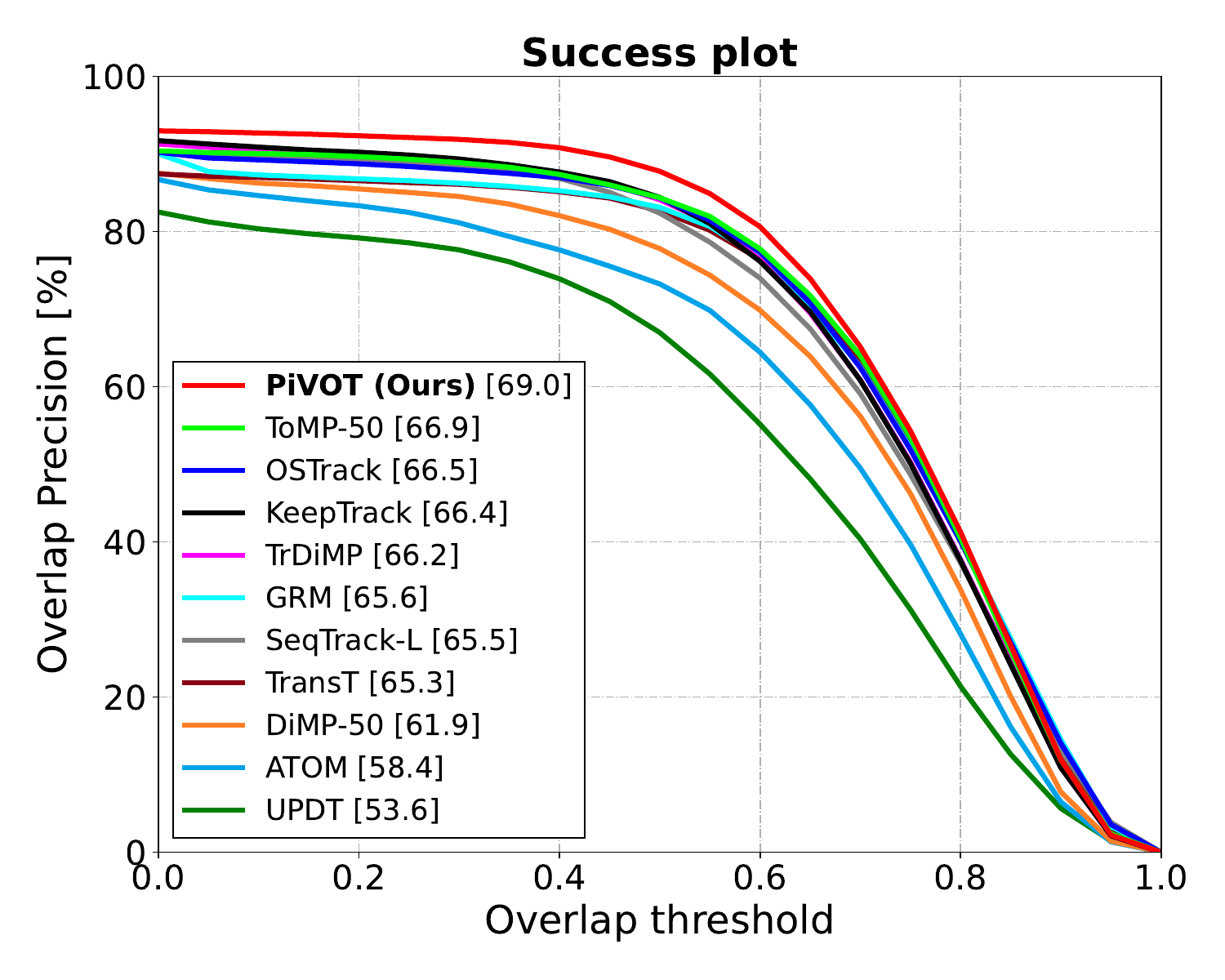}}}&
        
        \hspace{-0.3cm}
        
        {{\includegraphics[scale=0.22]
        {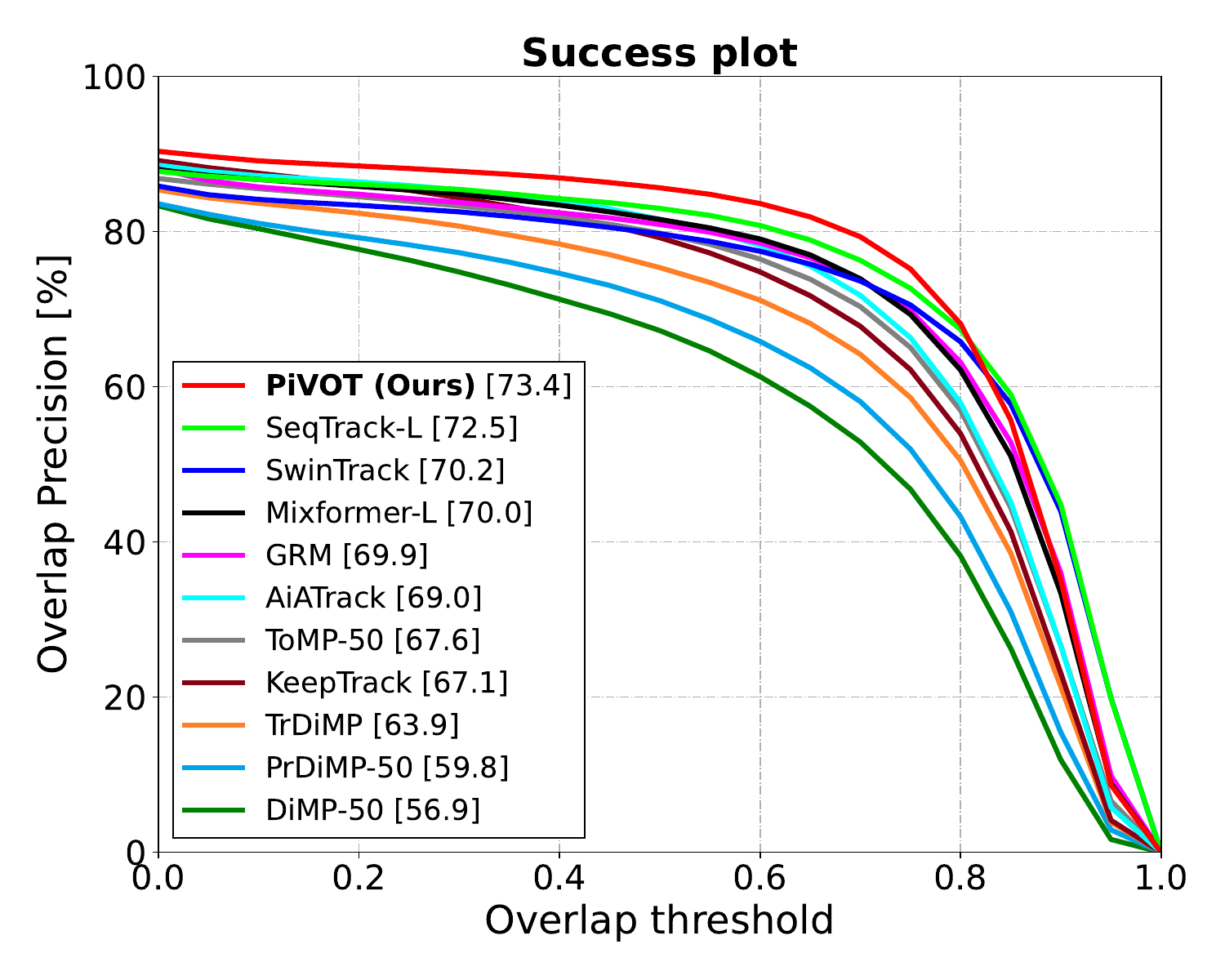}}}&

        \hspace{-0.3cm}
        
        {{\includegraphics[scale=0.22]
        {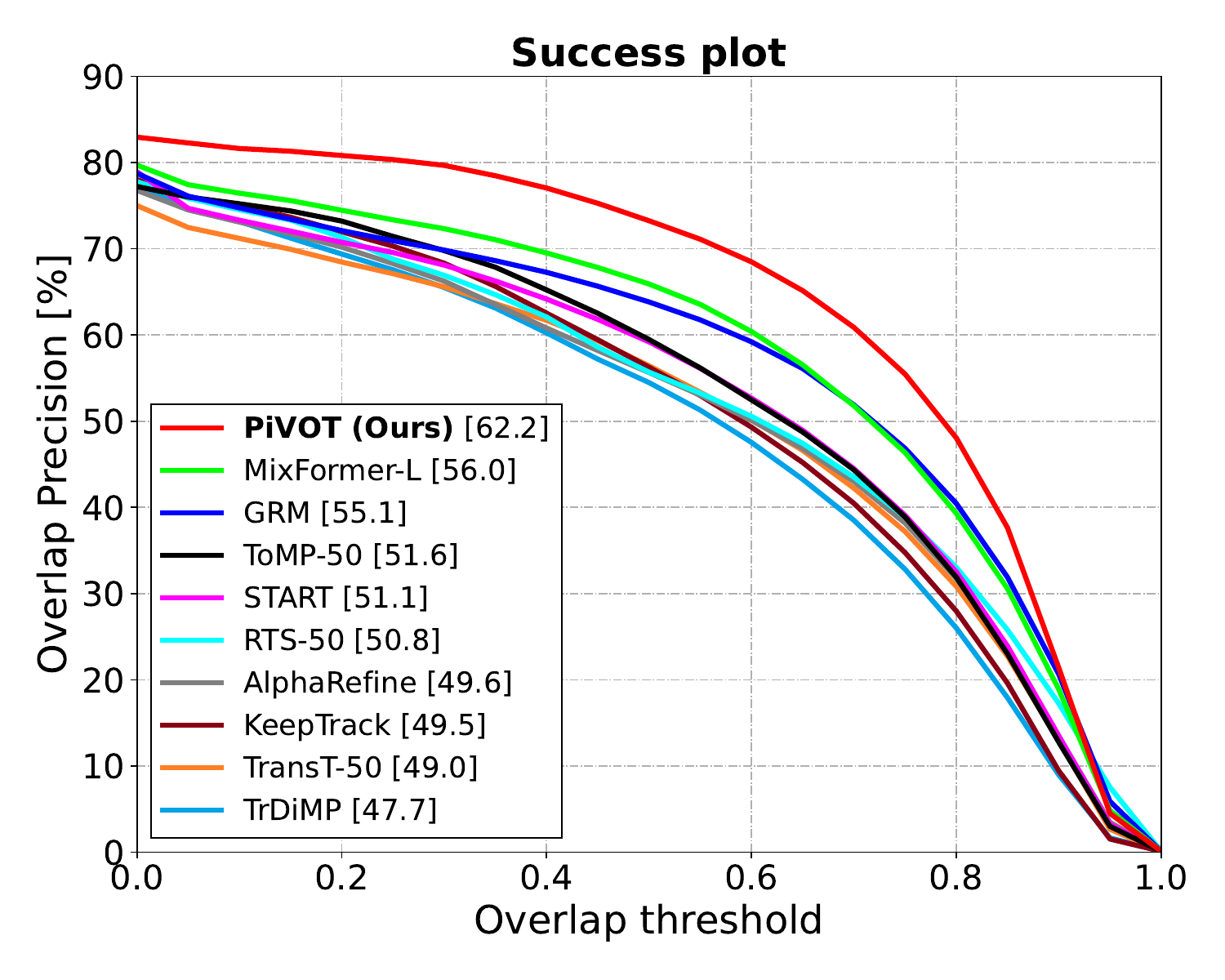}}}
        
        \\

        (a) NfS~\cite{NfS} & 
        (b) LaSOT~\cite{LaSOT} & 
        (c) AVisT~\cite{AVisT}

        \end{tabular}
        
 
	\caption{
        It shows the success plots of the proposed and competing methods on the NfS, LaSOT, and AVisT datasets with AUC scores in the legend.
        } 
	\label{fig:Success_plots_NfS_otb_uav}
 \vspace{-0.15in}
\end{figure*}

 
\section{Experiments}
\label{sec:exp}

Our method is evaluated in this section. 
We detail the implementation, training, and testing setups.
Next, we compare our method with state-of-the-art methods and analyze their performances.
Finally, we perform ablation studies to validate the contributions of individual components.

\textbf{Implementation Details.}
Our method was developed using PyTorch 1.10 and CUDA 11.3 for PiVOT-50 and PyTorch 2.0.0 with CUDA 11.7 for PiVOT-L, all within the PyTracking framework~\cite{pytracking}.
%
%
\revone{
We sample 200k sub-sequences and train the model for a total of 100 epochs using NVIDIA RTX 3090 GPUs.
AdamW~\cite{ADAMW} is used as the optimization solver.
There are two stages in the training process. 
For PiVOT-L, the backbone is frozen during both training stages since it leverages ViT-L/14 from DINOv2~\cite{DINOv2} as its backbone, a vision foundation model with strong generalization capability. 
PiVOT-50 uses the ResNet-50 backbone and is fine-tuned in the first stage but frozen in the second stage.
}

\revone{
Specifically, in the first stage, we train the tracker without the prompting components for 60 epochs,}
excluding the prompting components, for 60 epochs with a learning rate of \( 10^{-4} \). This rate decays by 0.2 after 30 and 50 epochs, producing a pre-trained tracking model. 
\revone{
Following this, in the second stage, we integrate}
the pre-trained tracking model and fine-tune our prompting components with the pre-trained tracking model for an additional 40 epochs.
The learning rate for the prompting components is initiated as \( 5 \times 10^{-3}\), while the learning rate for fine-tuning the pre-trained tracking model is set to \( 4 \times 10^{-6}\). 
This low value ensures that the pre-trained model maintains its discriminative capability while adapting to the refined features from the prompting components. The learning rates decay in the last 10 epochs.
The difference between training PiVOT-50 and PiVOT-L depends mainly on the backbones they adopt.
We set $\lambda_{reg}$ = 1, $\lambda_{cls}$ = 100, and $\lambda_{can}$ = 10
in the experiments. 
For prompt refinement, we use the official ViT-L/14@336px CLIP model~\cite{CLIP}.

\vspace{+0.1in}

\textbf{Training and Inference Setup.}
We adopt the training splits from the LaSOT~\cite{LaSOT}, GOT10k~\cite{GOT-10k}, TrackingNet~\cite{TrackingNet}, and MS-COCO~\cite{MS-COCO} datasets for model training. 
A training sub-sequence for each batch is constructed by randomly sampling two training frames and a test frame within a 200-frame window in a video. 
Image patches are extracted after random translation and scaling relative to the bounding box of the target object. 
Random flipping and color jittering are applied for augmentation.
Following ToMP, we set the image resolution and search area scale factor to 288$\times$288 and 5.0, respectively.
In PiVOT-L-22, the image resolution 
is  
\(378 \times 378\). The output ViT patch tokens are reshaped, and the resolution is reduced from \(27 \times 27\) to \(22 \times 22\) using adaptive average pooling for efficient training. 
In PiVOT-L-27, the output feature resolution remains \(27 \times 27\) without pooling, and 
all other settings are identical to those in PiVOT-L-22.
The difference between PiVOT-50 and PiVOT-L lies in the backbones they adopt.
Both employ a single-layer adapter for GOT adaptation.

During testing, we evaluate our proposed 
PiVOT on eight benchmarks, including the 
OTB-100~\cite{OTB}, 
UAV123~\cite{UAV},
NfS~\cite{NfS}, 
LaSOT~\cite{LaSOT},
TrackingNet~\cite{TrackingNet},
GOT-10k~\cite{GOT-10k} and
AVisT~\cite{AVisT}
datasets as well as VOT2022~\cite{VOT2022} challenge.
We set the confidence threshold $\tau$ to 0.05
and $\gamma$ is 0.25.
We follow \cite{KeepTrack} to set most of the hyper-parameters in Eq.~\ref{eqn:can_extract}, as our method shares a similar idea for candidate extraction.
Our tracker is evaluated on a single NVIDIA RTX 3090 GPU, 
with approximately 4GB of GPU memory usage during evaluation.
\vspace{-0.30in}


\subsection{\textbf{Comparisons with the State-of-the-Art Methods}}


\begin{table}[]
\caption{
Comparing our method and the competing methods on 
multiple datasets 
using Success and Precision AUC.
}
\resizebox{0.46\textwidth}{!}{
\begin{tabular}{ccccccccc}
\hline
 &  &  & \multicolumn{2}{c}{NfS~\cite{NfS}} & \multicolumn{2}{c}{OTB-100~\cite{OTB}} & \multicolumn{2}{c}{UAV123~\cite{UAV}} \\
\multirow{-2}{*}{Tracker} & \multirow{-2}{*}{Venue} & \multirow{-2}{*}{Backbone} & Suc & Pr & Suc & Pr & Suc & Pr \\ \hline
HCAT~\cite{HCAT}              & ECCV22    & ConvNet & 63.5 & -             & 68.1 & -    & 62.7          & -             \\
AiATrack~\cite{AiATrack}          & ECCV22    & ConvNet & 
{\color[HTML]{3166FF} \textbf{67.9}} & - & 69.6 & {\color[HTML]{3166FF} \textbf{91.7}} & {\color[HTML]{FE0000} \textbf{70.6}} & {\color[HTML]{FE0000} \textbf{90.7}}      
\\
ToMP-50~\cite{ToMP}              & CVPR22    & ConvNet & 
66.9 & {\color[HTML]{3166FF} \textbf{80.6}} & 70.1 & 90.8 & 69.0 & 89.7   
\\
ToMP-101~\cite{ToMP}              & CVPR22    & ConvNet& 66.7 & 79.8          & 70.1 & 90.6 & 66.9          & 85.2         \\
CSWinTT~\cite{CSWinTT}           & CVPR22    & ConvNet & -    & -             & 67.1 & 87.2 & {\color[HTML]{3166FF} \textbf{70.5}}          & 90.3          \\
SwinTrack~\cite{SwinTrack} & NeurIPS22 & ConvNet & - & - & 69.1 & 90.2 & 69.8 & 89.1 \\ 
UCIF~\cite{UCIF} & TMM23 & ConvNet& 66.8 & 81.7 & 69.9 & 91.6 & 67.0 & -  \\
DATransT~\cite{DATransT} & TMM23 & ConvNet& 66.9 & - & {\color[HTML]{3166FF} \textbf{70.8}} & - & 69.7 \\
STRtrack~\cite{STRtrack} & IJCV23 & ConvNet& 66.9 & 79.9 & 70.7 & 91.0 & 69.6 & 88.6 \\
HSET~\cite{HSET} & TMM23 & ConvNet& - & - & 69.8 & {\color[HTML]{3166FF} \textbf{91.7}} & 54.4 & 76.2\\
\hline
\textbf{PiVOT-50} & - & ConvNet 
& 
{\color[HTML]{FF0000} \textbf{68.5}} & {\color[HTML]{FF0000} \textbf{82.6}} & {\color[HTML]{FF0000} \textbf{71.2}} & {\color[HTML]{FF0000} \textbf{92.3}} & 69.9 & {\color[HTML]{FF0000} \textbf{90.7}}
 \\ 
\hline \toprule
MixFormer-L~\cite{MixFormer}       & CVPR22    & ViT       & -    & -             & 70.4 & 92.2 & 69.5          & 90.9       \\
OSTrack-384~\cite{OSTrack} & ECCV22 & ViT & 
66.5 & 81.9 & 68.1 & 88.7 & {\color[HTML]{000000} 70.7} & {\color[HTML]{3166FF} \textbf{92.3}}
 \\
SeqTrack-L~\cite{SeqTrack}        & CVPR23    & ViT     & 65.5 & 81.9          & 68.3 & 89.1 & 68.5          & 89.1        \\
ARTrack-384~\cite{ARTrack} & CVPR23 & ViT & 66.8 & - & - & - & 70.5 & -  \\
GRM~\cite{GRM}               & CVPR23    & ViT     & 65.6 & 79.9          & 68.9 & 90.0 & 70.2          & 89.8               \\ 
CiteTracker~\cite{CiteTracker}        & ICCV23    & ViT     & -    & -             & 69.6 & 92.2 & -             & -           \\
F-BDMTrack~\cite{F-BDMTrack}         & ICCV23    & ViT     & 66.0 & -             & 69.5 & -    & 69.0          & -           \\
UVLTrack-L~\cite{UVLTrack} & AAAI24 & ViT & 67.6 & - & - & - & {\color[HTML]{FF0000} \textbf{71.0}} & -  \\
\hline
\textbf{PiVOT-L-22} & - & ViT & {\color[HTML]{FF0000} \textbf{69.0}} & {\color[HTML]{FF0000} \textbf{85.6}} & {\color[HTML]{FF0000} \textbf{71.3}} & {\color[HTML]{3166FF} \textbf{94.1}} & {\color[HTML]{3166FF} \textbf{70.9}} & {\color[HTML]{FF0000} \textbf{92.8}} \\
\textbf{PiVOT-L-27} & - & ViT & {\color[HTML]{3166FF} \textbf{68.2}} & {\color[HTML]{3166FF} \textbf{84.5}} & {\color[HTML]{3166FF} \textbf{71.2}} & {\color[HTML]{FF0000} \textbf{94.6}} & {\color[HTML]{000000} 70.7} & {\color[HTML]{343434} 91.8}  \\ 
\hline \toprule
\end{tabular}
}
\label{tab:NFS_OTB_UAV_AUC_P}
\vspace{-0.25in}
\end{table}


\label{s4:SOTA_compare}
Like previous methods~\cite{ToMP, DiMP}, 
We evaluate PiVOT 
with 
success (Suc), precision (Pr), and normalized precision (NPr) AUC scores.
The precision score measures the center location distance between the predicted and ground truth targets, while the success score calculates their Intersection over Union (IoU). 
Detailed metric descriptions can be found in the appendix of our supplementary material.
To ensure consistency, we recalculated these metrics for all trackers using their raw predictions when available or the results reported in their papers. If both are missing, 
we refer to the survey paper~\cite{kugarajeevan2023transformers}. 
In the absence of data, we omit reporting the results.

Performance comparisons for NfS, OTB-100, and UAV123 are in Table~\ref{tab:NFS_OTB_UAV_AUC_P}. LaSOT results are detailed in Table~\ref{tab:lasot_auc_np_p}, with GOT-10K and TrackingNet in Table~\ref{tab: GOT_TrackingNet}. AVisT and VOT2022 performances are in Tables~\ref{tab: AVisT_auc_op50_op75} and \ref{tab: VOT22}, respectively. Further evaluations and comparisons are shown in Figure~\ref{fig:Success_plots_NfS_otb_uav} using success AUC plots. 
The methods include multiple trackers
~\cite{RTS,TrDiMP,TransT,KeepTrack,PrDiMP,Ocean,
SiamBAN,SiamMask,DaSiamRPN,SiamRPN++,ATOM,UPDT
}.
Please note that plot presentation requires the raw result of the tracker. We will not include the results if the method does not provide the raw result or the pre-trained model.

Our method generalizes well on diverse datasets. 
We discuss the performance across the benchmarks as follows:

\textbf{NfS}~\cite{NfS}: 
We present results from the 30 FPS version Need for Speed (NfS) dataset, 
designed for testing without a training set, consisting of 100 short sequences.
Figure~\ref{fig:Success_plots_NfS_otb_uav}(a) and Table~\ref{tab:NFS_OTB_UAV_AUC_P} display the success plot, success AUC and precision AUC, respectively.
PiVOT-50 outperforms ToMP-50 by 2\% in precision score,
surpassing trackers that use the ConvNet (CNN) backbone. %
PiVOT-L outperforms transformer-based trackers, setting new records in success and precision AUC scores.

\textbf{OTB-100}~\cite{OTB}: 
This short-sequence dataset, designed solely for testing without a training set, consists of 100 sequences.
Table~\ref{tab:NFS_OTB_UAV_AUC_P} displays the AUC scores.
PiVOT-50 outperforms ToMP-50 by 1.5\% in precision AUC. 
PiVOT-L outperforms trackers that use the transformer backbone and has set new records in both success and precision AUC scores.
%



\begin{table}[]
\caption{
Comparison of our method and competitors on \textbf{\textit{LaSOT}}~\cite{LaSOT}. 
}
\resizebox{0.48\textwidth}{!}{
\begin{tabular}{cccccc}
\hline
Tracker & Venue & Backbone & Suc & NPr & Pr \\ \hline
STARK~\cite{STARK}        & ICCV21 & ConvNet   & 66.4          & 76.3           & 71.2            \\
AutoMatch~\cite{AutoMatch}         & ICCV21 & ConvNet   & 58.3          & -              & 59.9   \\
HCAT~\cite{HCAT}              & ECCV22 & ConvNet   & 59.3          & 68.7           & 61.0           \\
CIA~\cite{CIA}               & ECCV22 & ConvNet   & 66.2          & -              & 69.6                \\
ToMP-50~\cite{ToMP}           & CVPR22 & ConvNet   & 67.6          & {\color[HTML]{3166FF} \textbf{78.0}}    & 72.2    \\
UTT~\cite{UTT}               & CVPR22 & ConvNet   & 64.6          & -              & 67.2           \\
CSWinTT~\cite{CSWinTT}           & CVPR22 & ConvNet   & 
66.2 & 75.2 & 70.9   \\
GTELT~\cite{GTELT}              & CVPR22    & ConvNet   & {\color[HTML]{3166FF} \textbf{67.7}} & 75.9 & {\color[HTML]{FF0000} \textbf{73.2}}       \\
GdaTFT~\cite{GdaTFT}            & AAAI23 & ConvNet & 64.3          & 68.0           & 68.7       \\
DETA~\cite{DETA}                 & TMM23     & ConvNet   & 66.0                                 & 74.8                                 & 70.1                                  \\
DATransT~\cite{DATransT}             & TMM23     & ConvNet   & 65.2                                 & 73.6                                 & 69.3                                \\
HSET~\cite{HSET}                 & TMM23     & ConvNet   & 37.2                                 & -                                    & 35.4                                   \\ \hline
\textbf{PiVOT-50} & -      & ConvNet   & 
{\color[HTML]{FE0000} \textbf{68.3}} & {\color[HTML]{FF0000} \textbf{78.9}} & {\color[HTML]{3166FF} \textbf{73.1}}         
  \\ 
\hline \toprule
OSTrack-384~\cite{OSTrack}       & ECCV22 & ViT       & 71.1          & 81.1           & 77.6          \\
ZoomTrack~\cite{ZoomTrack}         & NeurIPS23 & ViT       & 70.2          & -              & 76.2            \\
TATrack~\cite{TATrack}           & AAAI23 & ViT       & 71.0          & 79.1           & 76.1  \\
CTTrack~\cite{CTTrack}            & AAAI23 & ViT       & 69.8          & 79.7           & 76.2          \\
VideoTrack~\cite{VideoTrack}        & CVPR23 & ViT    & 70.2          & -              & 76.4         \\
GRM~\cite{GRM}               & CVPR23 & ViT       & 69.9          & 79.3           & 75.8                \\ 
DropTrack~\cite{DropMAE} & CVPR23 & ViT & 71.5 & 81.5 & 77.9  \\
MAT\_freeze~\cite{MAT}               & CVPR23 & ViT      & 65.2          & 74.8           & -               \\
SeqTrack-L~\cite{SeqTrack} & CVPR23 & ViT & 72.5 & 81.5 & 79.2 \\
ARTrack-384~\cite{ARTrack} & CVPR23 & ViT & 72.6 & 81.7 & 79.1\\
ROMTrack-384~\cite{ROMTrack}          & ICCV23 & ViT    & 71.4 & 81.4                                 & 78.2      \\
CiteTracker~\cite{CiteTracker}       & ICCV23 & ViT       & 69.7          & 78.6           & 75.7         \\
F-BDMTrack~\cite{F-BDMTrack}        & ICCV23 & ViT      & 69.9          & 79.4           & 75.8         \\
EVPTrack-384~\cite{EVPTrack} & AAAI24 & ViT & {\color[HTML]{3166FF} \textbf{72.7}} & 82.9 & {\color[HTML]{3166FF} \textbf{80.3}} \\
UVLTrack-L~\cite{UVLTrack} & AAAI24 & ViT & 71.3 & - & 78.3\\
Linker-384~\cite{Linker}          & TMM24     & ViT       & 71.5                                 & 81.2                                 & 78.1                                   \\
AQATrack~\cite{AQATrack} & CVPR24 & ViT & {\color[HTML]{3166FF} \textbf{72.7}} & 82.9 & 80.2 \\
DiffusionTrack-L~\cite{DiffusionTrack} & CVPR24 & ViT & 72.3 & 81.8 & 79.1\\
HIPTrack~\cite{HIPTrack} & CVPR24 & ViT & {\color[HTML]{3166FF} \textbf{72.7}} & 82.9 & 79.5 \\
OneTracker~\cite{OneTracker} & CVPR24 & ViT & {\color[HTML]{000000} 70.5} & 79.9 & 76.5  \\
\hline
\textbf{PiVOT-L-22} & - & ViT & {\color[HTML]{343434} 71.8} & {\color[HTML]{3166FF} \textbf{83.6}} & 80.1 \\
\textbf{PiVOT-L-27} & - & ViT & {\color[HTML]{FE0000} \textbf{73.4}} & {\color[HTML]{FE0000} \textbf{84.7}} & {\color[HTML]{FE0000} \textbf{82.1}}
\\ 
\hline \toprule
\end{tabular}
}
\vspace{-0.05in}
\label{tab:lasot_auc_np_p}
\end{table}


\begin{figure}[!t] \centering
\includegraphics[width=0.48\textwidth]{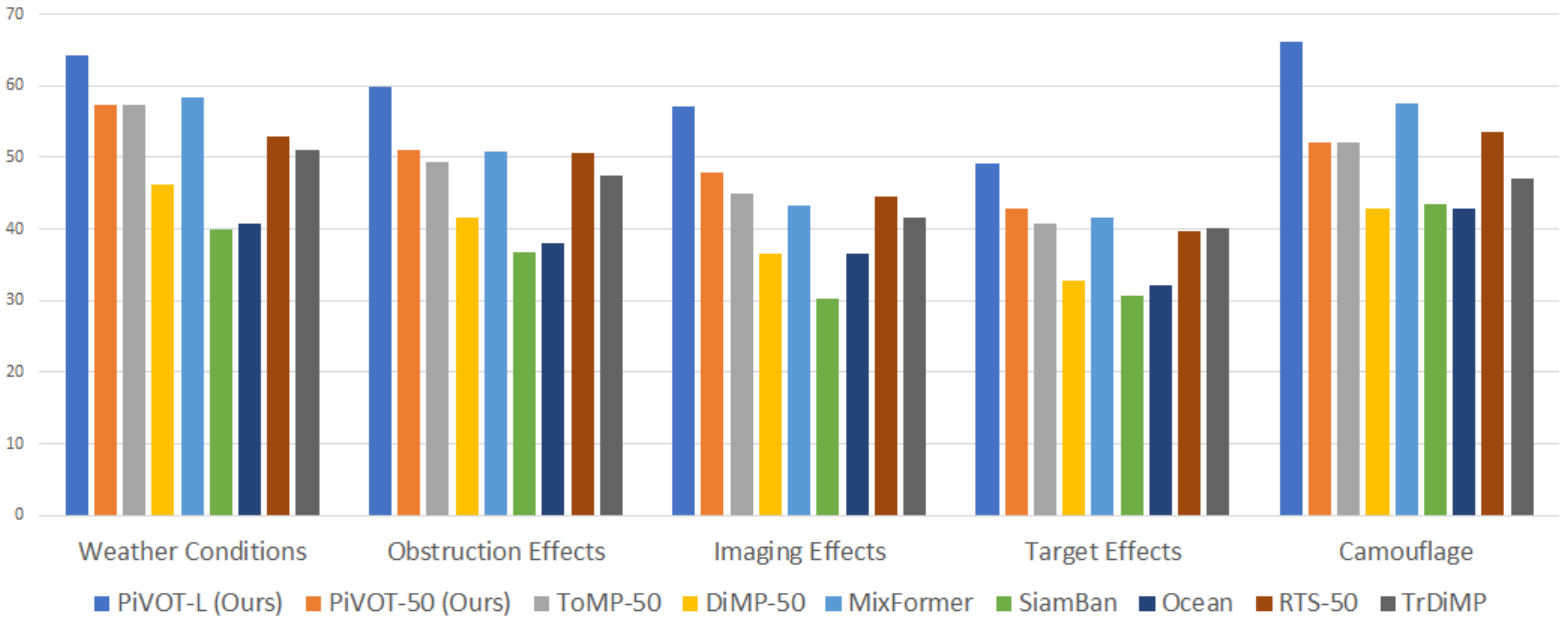}
\caption{
Attribute analysis on AVisT compares PiVOT with multiple trackers.
} \label{fig:per_attr_ana}
\vspace{-0.15in}
\end{figure}

\textbf{UAV123}~\cite{UAV}:
The dataset includes 123 short sequences without the corresponding training set. Around 90\% of the targets are persons and cars observed from a UAV perspective. 
Table~\ref{tab:NFS_OTB_UAV_AUC_P} displays the success AUC and precision AUC scores.
PiVOT sets a new record among trackers.
\vspace{0.1in}

\textbf{LaSOT}~\cite{LaSOT}:
This dataset has 280 long-term sequences with overlapping object classes in training and test sets. 
PiVOT-50 outperforms trackers with a CNN backbone in terms of normalized precision AUC scores.
PiVOT-L has set new records in success, precision and normalized precision AUC scores.
Figure~\ref{fig:Success_plots_NfS_otb_uav}(b) and Table~\ref{tab:lasot_auc_np_p} display the success plot, success AUC and precision AUC, respectively.

\textbf{AVisT}~\cite{AVisT}:
This new benchmark, designed for testing without a training set, covers 120 short and long-duration sequences. 
Figure~\ref{fig:Success_plots_NfS_otb_uav}(c) and Table~\ref{tab: AVisT_auc_op50_op75} display the success plot, success AUC, OP50, and OP75, respectively.
PiVOT-50 outperforms trackers with a CNN backbone, 
PiVOT-L achieves the state-of-the-art performance.
Attribute analysis in Figure~\ref{fig:per_attr_ana} highlights how our method performs better than the baseline in scenarios such as Target Effects (e.g., distractor or deformation object), Obstruction Effects (e.g., occlusion), and Imaging Effects (e.g., images with noise).
\vspace{0.05in}

\textbf{GOT-10k}~\cite{GOT-10k}:
This dataset comprises 420 short-term sequences featuring non-overlapping object classes in the training and test sets. Adhering to the official requirements, we exclusively utilize the training split of GOT-10k for training purposes in evaluating this benchmark.
Table~\ref{tab: GOT_TrackingNet} displays the average overlap (AO) and success rates (SR).
Our PiVOT exhibits strong performance in this benchmark.

\textbf{TrackingNet}~\cite{TrackingNet}:
This dataset has 511 short-term sequences with overlapping object classes in training and test sets. 
Table~\ref{tab: GOT_TrackingNet} displays the success AUC and precision AUC scores.
SeqTrack-L performs well here. 
PiVOT-L ranked as runner-up in Normalized Precision and Precision AUC scores.
%


\textbf{VOT2022}~\cite{VOT2022}:
We evaluate the 2022 edition of the Visual Object Tracking short-term challenge. 
Table~\ref{tab: VOT22} presents a series of evaluated methods.
Our PiVOT achieves the highest robustness score. 
We use our PiVOT-L-22 and benchmark it against the best-performing models of other methods in this comparison.
Note that the VOT challenge allows the use of additional training data.
We evaluate the method using the same training data and the trained model as described in our paper, with the setting being the same as that of ToMP (tomp).

Overall, PiVOT performs well on datasets with object classes that are out of the distribution in the training data, such as NfS and AVisT. 
When evaluated on in-distribution datasets like LaSOT and TrackingNet, it achieves comparable results among trackers on the 
Suc score and excels in the NPr score.

\vspace{-0.10in}

\begin{table}[]
\caption{
Comparisons of our method with the competing methods on \textbf{\textit{AVisT}}~\cite{AVisT} using multiple evaluated metrics. 
All results of compared methods are directly cited from their papers if available, or from the results reported in the AVisT paper.
}
\vspace{-0.1in}
\resizebox{0.46\textwidth}{!}{

\begin{tabular}{cccccc}
\hline
Tracker        & Venue  & Backbone & Suc & OP50 & OP75 \\ \hline
TransT~\cite{TransT}         & CVPR21 & ConvNet& 49.0          & 56.4 & 37.2 \\
TrDiMP~\cite{TrDiMP}         & CVPR21 & ConvNet& 48.1          & 55.3 & 33.8 \\
TrSiam~\cite{TrDiMP}         & CVPR21 & ConvNet& 47.8          & 54.8 & 33.0 \\
AlphaRefine~\cite{AlphaRefine}    & CVPR21 & ConvNet& 49.6          & 55.6 & 38.2 \\
STARK~\cite{STARK}          & ICCV21 & ConvNet& 
51.1 & 59.2 & {\color[HTML]{3166FF} \textbf{39.1}}
\\
KeepTrack~\cite{KeepTrack}      & ICCV21 & ConvNet& 49.4          & 56.3 & 37.8 \\
ToMP-50~\cite{ToMP}           & CVPR22 & ConvNet& 
{\color[HTML]{3166FF} \textbf{51.6}} & {\color[HTML]{3166FF} \textbf{59.5}} & 38.9 
\\
\hline
\textbf{PiVOT-50}   & - & ConvNet& 
{\color[HTML]{FF0000} \textbf{52.5}} & {\color[HTML]{FF0000} \textbf{60.7}} & {\color[HTML]{FF0000} \textbf{39.2}}                    
\\ 
\hline \toprule
MixFormer-22k~\cite{MixFormer}  & CVPR22 & ViT      & 53.7          & 63.0 & 43.0 \\
MixFormerL-22k~\cite{MixFormer} & CVPR22 & ViT      & 56.0          & 65.9 & 46.3 \\
GRM~\cite{GRM}            & CVPR23 & ViT    & 54.5          & 63.1 & 45.2 \\ 
UVLTrack-L~\cite{UVLTrack} & AAAI24 & ViT & 57.8 & 67.9 & 48.7 \\
\hline
\textbf{PiVOT-L-22} & - & ViT & {\color[HTML]{3166FF} \textbf{61.2}} & {\color[HTML]{3166FF} \textbf{72.8}} & {\color[HTML]{3166FF} \textbf{54.1}} \\
\textbf{PiVOT-L-27} & - & ViT & {\color[HTML]{FE0000} \textbf{62.2}} & {\color[HTML]{FE0000} \textbf{73.3}} & {\color[HTML]{FE0000} \textbf{55.5}} \\ 
\hline \toprule
\end{tabular}

}
\vspace{-0.05in}
\label{tab: AVisT_auc_op50_op75}
\end{table}


\begin{table}[]
\caption{
Comparisons of our method with the competing methods on 
\textbf{\textit{GOT-10k}}~\cite{GOT-10k} and \textbf{\textit{TrackingNet}}~\cite{TrackingNet}.
}
\vspace{-0.1in}
\resizebox{0.46\textwidth}{!}{

\begin{tabular}{cccccccc}
\hline
Tracker & \multirow{2}{*}{Venue} & \multicolumn{3}{c}{GOT-10k} & \multicolumn{3}{c}{TrackingNet} \\
 &  & AO & SR(0.50) & SR(0.75) & Suc & NPr & Pr \\ \hline
MixFormer-L~\cite{MixFormer} & CVPR22 & 70.7 & 80.0 & 67.8 & 83.9 & 88.9 & 83.1 \\
CTTrack-L~\cite{CTTrack} & AAAI23 & 72.8 & 81.3 & 71.5 & 84.9 & 89.1 & 83.5 \\
SeqTrack-L~\cite{SeqTrack} & CVPR23 & 74.8 & 81.9 & 72.2 & \textbf{85.5} & 89.8 & \textbf{85.8} \\
ROMTrack-384~\cite{ROMTrack} & ICCV23 & 74.2 & 84.3 & 72.4 & 84.1 & 89.0 & 83.7 \\
ZoomTrack~\cite{ZoomTrack} & NeurIPS23 & 73.5 & 83.6 & 70.0 & 83.2 & - & 82.2 \\ 
UVLTrack-L~\cite{UVLTrack}  & AAAI24 & - & - & - & 84.1 & - & 82.9 \\
OneTracker~\cite{OneTracker} & CVPR24 & - & - & - & 83.7 & 88.4 & 82.7 \\
\hline
\textbf{PiVOT-L-22} & - & 75.9 & 87.5 & 74.2 & 84.3 & 89.2 & 83.9 \\
\textbf{PiVOT-L-27} & - & \textbf{76.9} & \textbf{87.6} & \textbf{75.5} & 85.3 & \textbf{90.0} & 85.3 \\ \hline
\end{tabular}
}
\label{tab: GOT_TrackingNet}
\end{table}



\begin{table}[!t]
\caption{
Comparisons of our method with the competing methods on 
\textbf{\textit{VOT2022}}~\cite{VOT2022}.
}
\vspace{-0.1in}
\resizebox{0.46\textwidth}{!}{
\begin{tabular}{ccccccc}
\hline
\multirow{2}{*}{Tracker} & \multirow{2}{*}{\textbf{PiVOT}} & \multirow{2}{*}{MixFormerL} & \multirow{2}{*}{OSTrackSTB} & \multirow{2}{*}{TransT\_M} & \multirow{2}{*}{SwinTrack} & \multirow{2}{*}{tomp} \\
                         &                                 &                             &                             &                            &                            &                       \\ \hline
EAO                      & 0.560                           & \textbf{0.602}              & 0.591                       & 0.537                      & 0.524                      & 0.511                 \\
Robustness               & \textbf{0.873}                  & 0.859                       & 0.869                       & 0.849                      & 0.803                      & 0.818                 \\ \hline
\end{tabular}

}
\label{tab: VOT22}
\end{table}



\begin{table}[]
\caption{
Ablation studies on the feature prompting in terms of precision AUC score.
%
\revone{
``Initial'' indicates the initial prompt, 
while ``Refined'' indicates the initial prompt after applying CLIP refinement.
The last row shows performance through the prompting mechanism with the CLIP-refined visual prompt.
}
}
\resizebox{0.46\textwidth}{!}{

\begin{tabular}{cccccccc}
\hline
Tracker & Initial & Refined & NfS & OTB-100 & UAV123 & AVisT& LaSOT \\ \hline
ToMP-50 &  &  & 80.6 & 90.8 & 89.7 & 47.7 & 72.2 \\
PiVOT-50 &  &  & 80.8 & 90.8 & 88.8 & 47.8 & 71.7 \\
PiVOT-50 & Y &  & 80.5 & 90.1 & 89.7 & 47.5 & 72.0 \\
PiVOT-50 & Y & Y & \textbf{82.6} & \textbf{92.3} & \textbf{90.7} & \textbf{48.6} & \textbf{73.1} \\ \hline
ToMP-L &  &  & 84.3 & 93.1 & 91.0 & 63.4 & 79.1 \\
PiVOT-L &  &  & 84.3 & 93.2 & 90.2 & 63.5 & 78.5 \\
PiVOT-L & Y &  & 84.1 & 92.8 & 91.1 & 63.0 & 79.0 \\
PiVOT-L & Y & Y & \textbf{85.6} & \textbf{94.1} & \textbf{92.8} & \textbf{64.5} & \textbf{80.1} \\ \hline
\end{tabular}

}
\label{tab: Ablation_for_feature_prompting_PR}
\vspace{-0.1in}
\end{table}


\subsection{\textbf{Ablation Studies}}
\vspace{0.05in}
\label{s4:ablation}
In the ablation studies, we assess if the prompt refined by CLIP~\cite{CLIP} enhances the tracker performance. 
Table~\ref{tab: Ablation_for_feature_prompting_PR} reports the results. 
For each backbone, the first row highlights the baseline method, ToMP, which serves as the pre-trained tracking model for our PiVOT. 
In the second row, PiVOT is introduced. Without using any visual prompt during inference, it leverages components identical to ToMP, delivering performance on par with ToMP.
Introducing an initial prompt for feature refinement without CLIP, the performance of PiVOT rises on in-distribution datasets like UAV123 and LaSOT but falls on out-of-distribution datasets like NfS, OTB-100, and AVisT, 
\revone{
The performance drops occur because prompting with only the initial prompt, without CLIP refinement, cannot properly handle unseen situations and yields suboptimal results,}
as seen in the third row.
Upon incorporating CLIP for prompt refinement, the tracker notably outperforms the baseline. This refinement notably improves performance for PiVOT.
%

\vspace{-0.1in}
\subsection{\textbf{Visualization}}
\begin{figure}[!t] \centering
\includegraphics[width=0.45\textwidth]{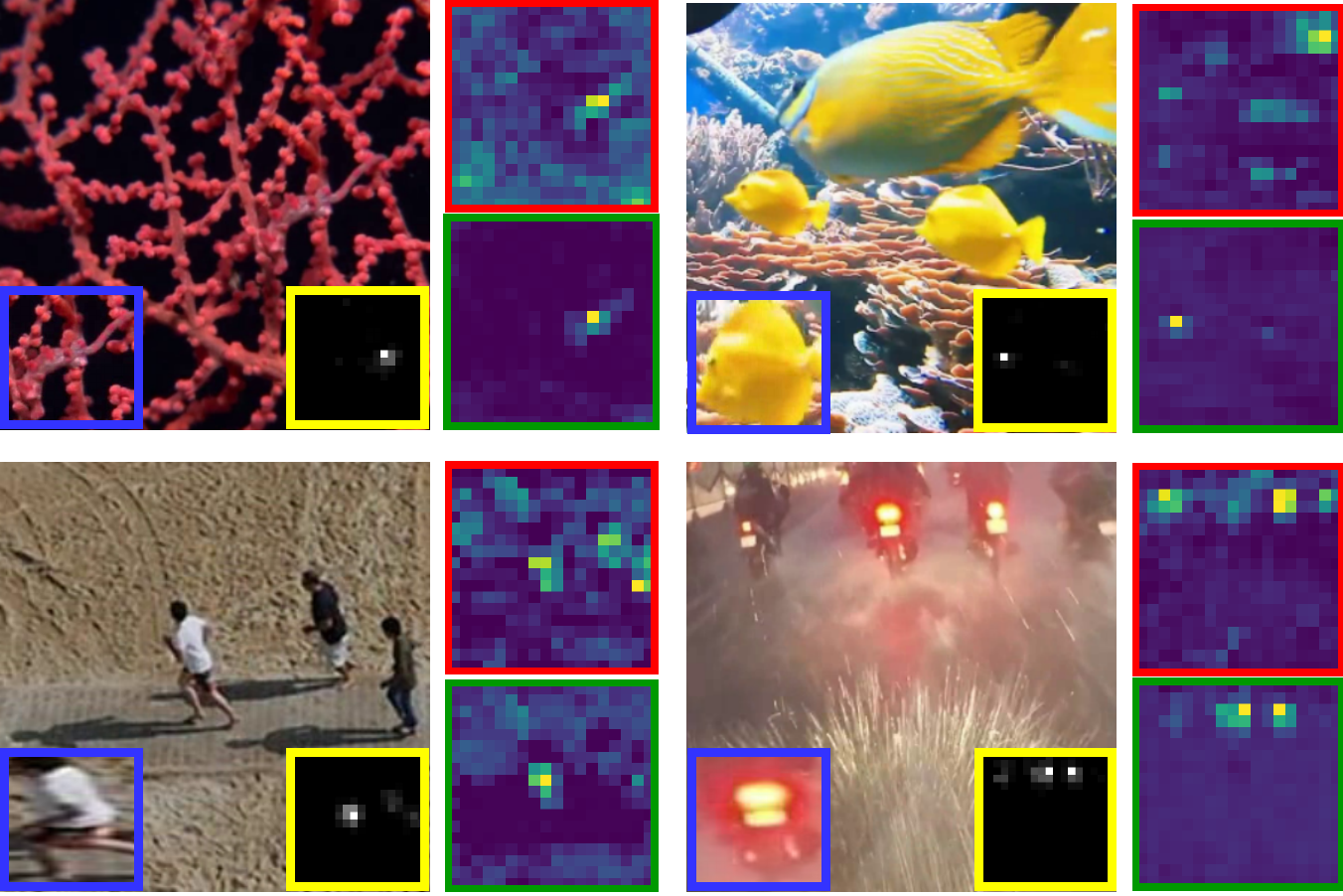}
\caption{
\textbf{Prompting visualisation.}
Given the current frame, we have a template in the blue box, a visual prompt in the yellow, a feature map in the red, and its prompted version after the RM application in the green. 
RM accentuates the visual prompt-highlighted area. 
We apply color mapping to the feature map to enhance visualization.
} \label{fig:prompting_vis}
\end{figure}

\textbf{Visual Prompting.}
We visualize examples of the prompting results in 
Figure~\ref{fig:prompting_vis}. 
It can be observed that the prompted feature maps emphasize the tracked objects and suppress most of the objects that the visual prompt does not highlight.
This is why more accurate tracking results are achieved through online CLIP knowledge transfer. 
%

\begin{figure*}[!t]
	\centering
	    \includegraphics[width=0.95\linewidth]{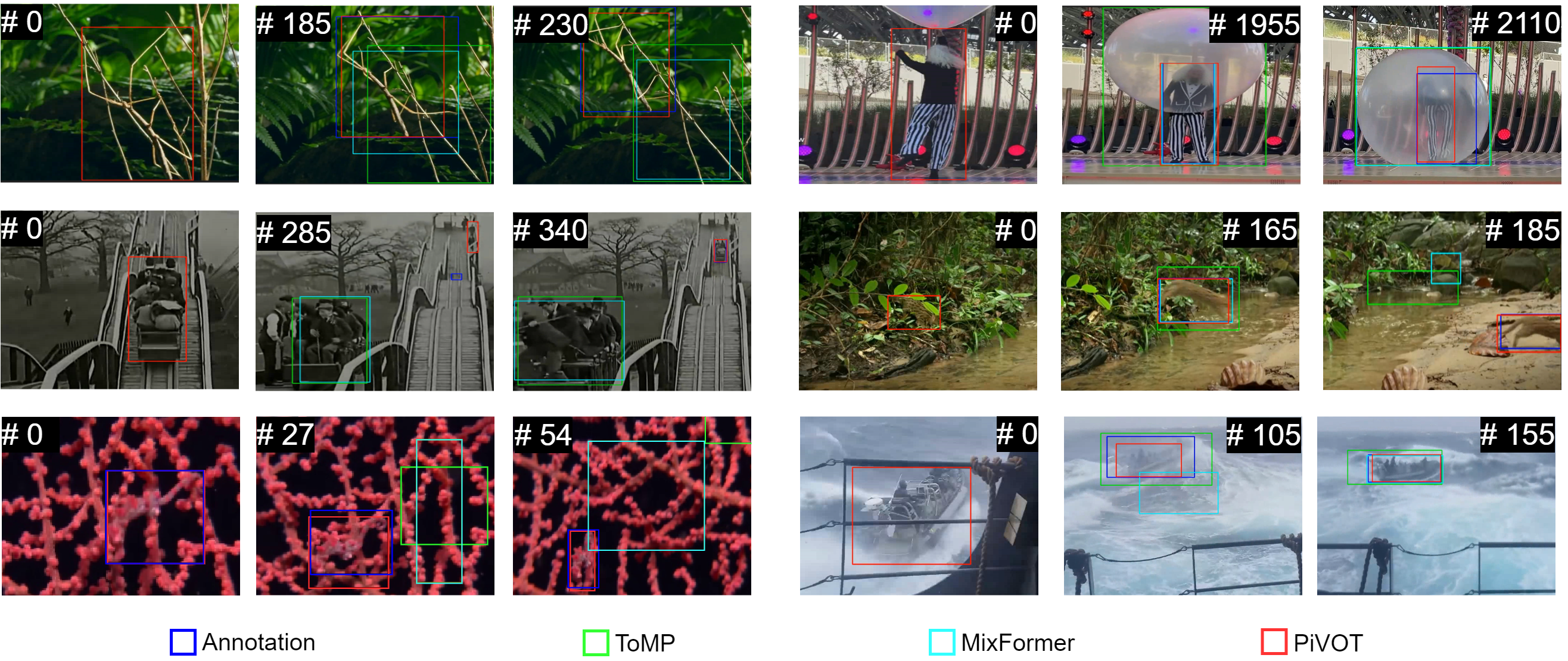}
     
    \vspace{0.05in}

	\caption{
        \textbf{Visual results.}
        Visual comparison of tracking results from different trackers (PiVOT, ToMP, and MixFormer) across various video sequences.
        }
	\label{fig: vis_2}
 \vspace{0.1in}
\end{figure*}

\noindent \textbf{
Visual Results among Trackers.}
We provide visual comparisons among trackers 
for more sequences in Figure~\ref{fig: vis_2}.
We can observe that our PiVOT is more discriminative than other trackers. 
Additionally, even when the tracker faces temporary occlusions leading to tracking failures, 
our tracker can still resume tracking after the occlusion recedes. 
This capability stems from our method employing the category prior with CLIP, 
which prevents the tracker from adapting to the wrong target and 
allows it to recover and track the initially identified category once the occlusion recedes 
(e.g., the roller-coaster case), 
thus showcasing the robustness of our method.
%
 
\begin{figure*}[!t]
	\centering
	    \includegraphics[width=0.95\linewidth]{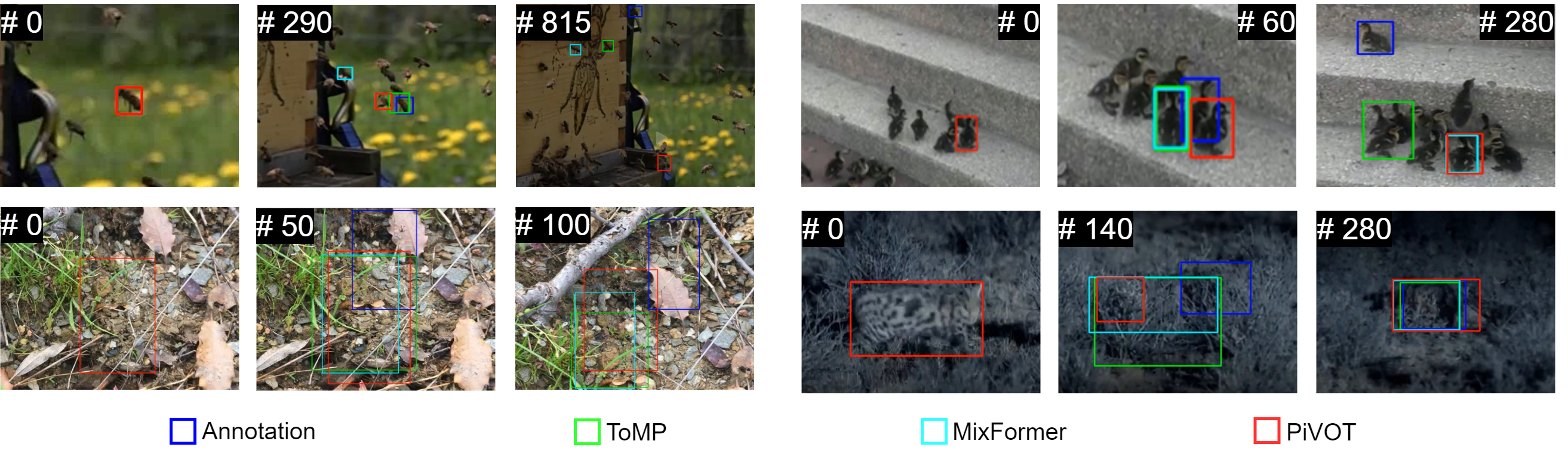}
     
    \vspace{-0.05in}
	\caption{
        \textbf{Failure cases.}
        Visual comparison of tracking results from different trackers (PiVOT, ToMP, and MixFormer) across various video sequences.
        %
        The primary challenges faced by our tracker include similar-looking distractors across frames (top row), limited semantic information (bottom-left), and occlusion (bottom-right). For more details, please refer to the text in the paper. 
        We provide 
        a video demo in the appendix.
        }
    \vspace{-0.05in}
	\label{fig: vis_3}
  \vspace{-0.05in}
\end{figure*}

\noindent \textbf{
Visual Illustration of Failure Cases.}
We also provide insights into the failure cases of our tracker, 
as illustrated in Figure~\ref{fig: vis_3}.
There are three major cases that our tracker struggles to handle effectively.
\textbf{First}, similar-looking distractors intertwine, as in scenarios of bees flying or ducklings jumping on stairs. 
Additionally, tracking becomes challenging when the target resolution is low. 
This challenge also encompasses limited semantic information and occlusion, 
which will be discussed in the following paragraphs.
\textbf{Second}, cases with limited semantic information can confuse the tracker. 
This is evident in the scenario shown in the left-middle of the figure (the stick insect case), 
where the bounding box contains more background region than the target itself, 
and the target closely resembles the background (camouflage).
\textbf{Third}, occlusion presents a challenge. 
As demonstrated in the 
bottom-right
of the figure, 
the tracker attempts to predict the wrong target when the target is occluded. 
Although the tracker may recover and resume tracking after the occlusion is removed, 
the ideal solution would be for the tracker to recognize the occlusion case and 
prevent adaptation to the wrong target during the occlusion. 
This will require further research and development.
%

\begin{table}[!t]
\caption{
Details of the PiVOT model variants were evaluated using the metrics of Precision (Pr) and Normalized Precision (NPr).
}

\resizebox{0.46\textwidth}{!}{
\begin{tabular}{ccccccccccc}
\hline
\multirow{2}{*}{Tracker} & \multicolumn{2}{c}{AVisT} & \multicolumn{2}{c}{LaSOT} & \multicolumn{2}{c}{NfS} & \multirow{2}{*}{\begin{tabular}[c]{@{}c@{}}Train\\ Mem\end{tabular}} & \multirow{2}{*}{\begin{tabular}[c]{@{}c@{}}Train\\ Batch\end{tabular}} & \multirow{2}{*}{\begin{tabular}[c]{@{}c@{}}Train\\ Param\end{tabular}} & \multirow{2}{*}{FPS} \\
 & Pr & NPr & Pr & NPr & Pr & NPr &  &  &  &  \\ \hline
\textbf{PiVOT-L-27} & \textbf{65.6} & \textbf{81.2} & \textbf{81.2} & \textbf{83.8} & 84.5 & 86.7 & $8\times24\,\text{GB}$ & 56 & 29M & 4 \\
\textbf{PiVOT-L-22} & 64.5 & 81.1 & 80.1 & 83.6 & \textbf{85.6} & \textbf{88.1} & $4\times24\,\text{GB}$  & 64 & 29M & 5 \\
MixFormer-L & 55.5 & 73.9 & 76.3 & 79.9 & - & - & $8\times32\,\text{GB}$ & 16 & 196M & 8 \\
SeqTrack-L & - & - & 79.2 & 81.5 & 81.9 & 84.4 & $8\times80\,\text{GB}$ & 8 & 309M & 5 \\ \hline
\end{tabular}
}
\label{tab: PiVOT_model_variants}
\vspace{-0.10in}
\end{table}


\vspace{-0.10in}
\subsection{\textbf{Computational Cost Analysis}}

\revone{
We 
provide an analysis of the computational costs for each component of our optimal model, PiVOT-L-27, as shown in Table~\ref{tab:PiVOT_cost}, 
detailing 
the time required to process a single video frame. 
This table measures the run time of each component in milliseconds, 
and presents the corresponding percentage of the running time of each component with respect to the total, where ``Adapter'' refers to the lightweight adapter attached to the backbone and 
``Head''  refers to the Tracking Head.
The primary computational bottleneck originates from the ``Backbone'', which utilizes the large-scale foundation model DiNOv2 with higher-resolution inputs (378$\times$378). Subsequently, the ``TPR'' module, 
a
significant model leveraging CLIP, requires high-resolution inputs (336$\times$336). The third one is the Tracking Head, which incorporates a multi-layer transformer architecture; however, it operates on a lower-resolution (27$\times$27) feature map, thus requiring lower computational demands compared to the aforementioned two components.
}


\vspace{-0.5in}
\subsection{\textbf{Limitations}}
As shown in Table~\ref{tab: PiVOT_model_variants},
although we have addressed the significant memory requirements compared to existing works that fine-tune 
vision transformer (ViT)
backbones during training, the inference 
remains 
a bottleneck for transformer-based methods. 
This is the case for our PiVOT, 
which requires inference from two foundation models: DiNOv2 and CLIP 
ViT
backbone during inference.
``TrainBatch'' indicates the usage of the trained batch size during training.
By freezing the backbone during training, we are able to use a larger batch size for training compared to other methods.
%
``TrainParam'' indicates the number of trainable parameters during training. Our PiVOT has 29M trainable parameters, including 22M for the Tracking Head and 7M newly introduced for PiVOT, 
resulting in only 9\% of the trainable parameters for tracker training compared to SeqTrack.
%
Both CLIP and DiNOv2 have nearly 300M parameters; 
however, their parameters are non-trainable during training in our PiVOT.
%
Adopting lightweight transformer tracking methods like HiT~\cite{HiT} 
or MixFormerV2~\cite{MixFormerV2}
, or adapting a lightweight foundation model, could enhance inference speed. However, this strategy potentially entails a compromise between speed and accuracy.



\begin{table}[!t]
\caption{Analysis of computational costs for each component of PiVOT.}
\resizebox{0.48\textwidth}{!}{
\begin{tabular}{cccccc|c}
Backbone & Adapter & PGN & TPR & RM & Head & Total (ms) \\ \hline
120.76 & 0.11 & 0.67 & 82.93 & 0.66 & 35.40 & 240.53 \\
50.21\% & 0.05\% & 0.28\% & 34.47\% & 0.27\% & 14.72\% & 100\%
\end{tabular}
}
\label{tab:PiVOT_cost}
\end{table}


\section{\revone{Attribute Analysis for GOT}}
\begin{figure*}[!ht]
	\centering
 
        \begin{tabular}{ccc}

        \hspace{+0.5cm}
        
        {{\includegraphics[scale=0.40]{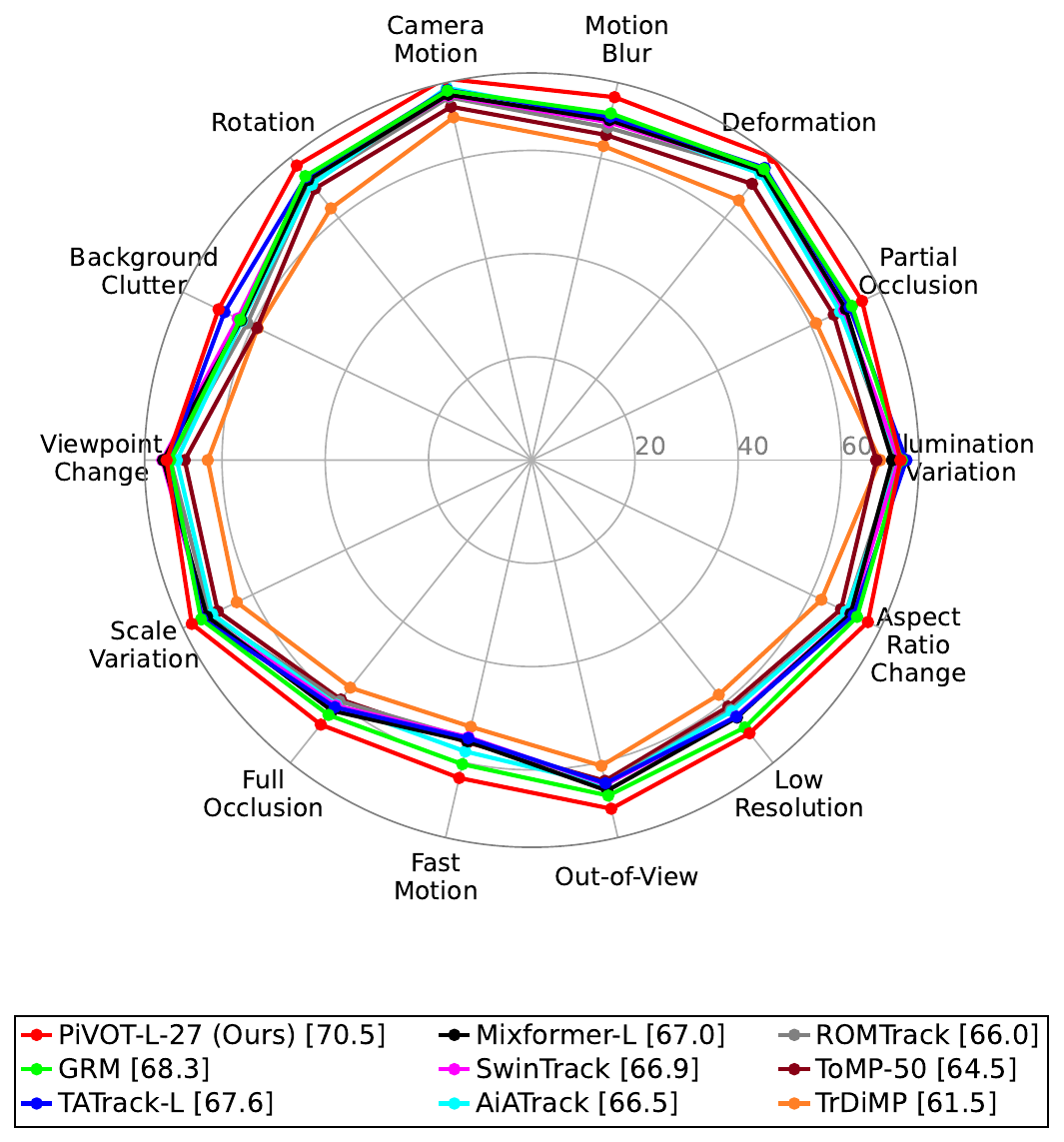}}}&
        
        
        {{\includegraphics[scale=0.40]{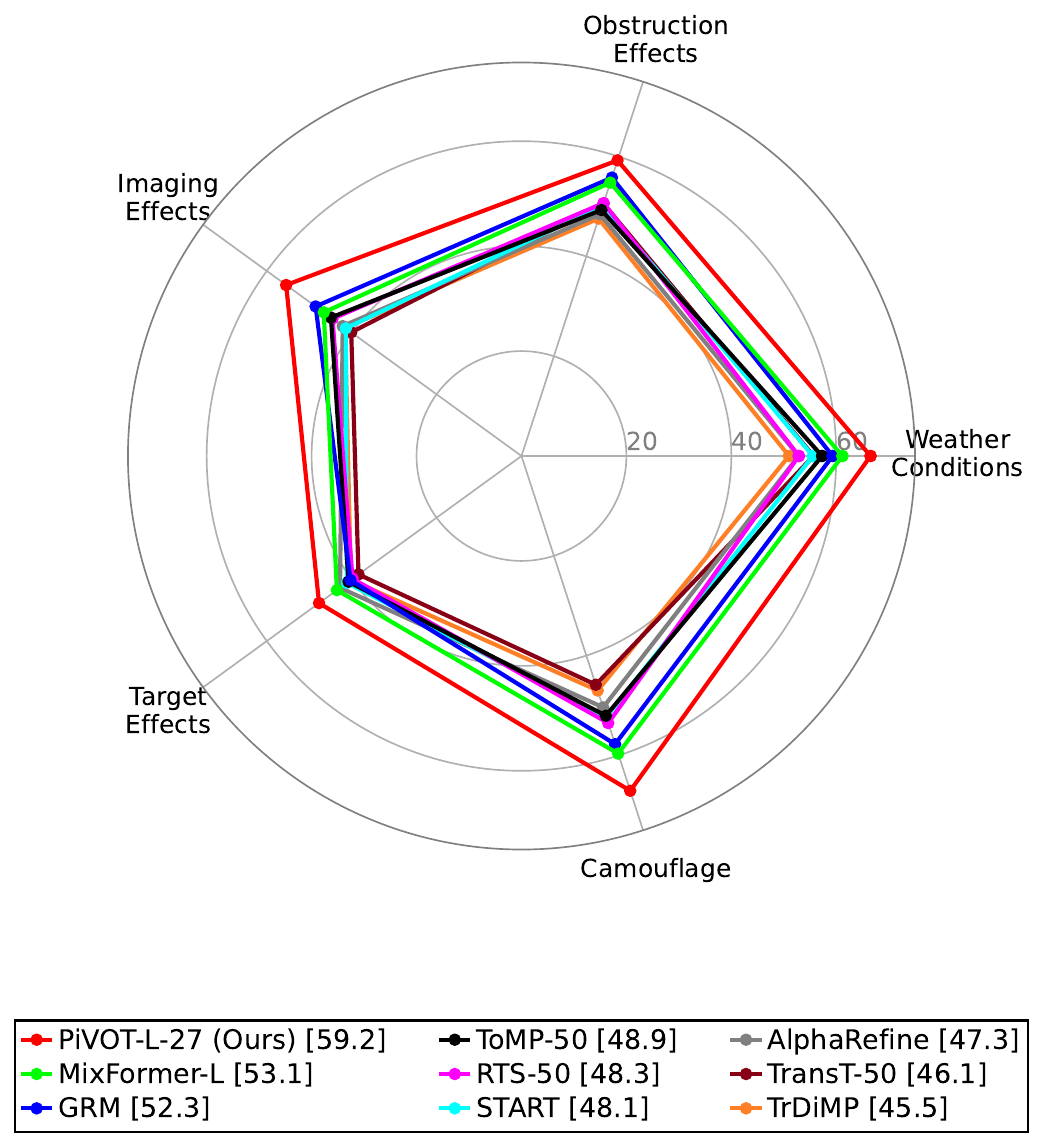}}}&

        \\

        (a) LaSOT & 
        (b) AVisT& 

        \end{tabular}
        
 
	\caption{
        \textbf{
        Attribute-based analysis of LaSOT and AVisT, comparing PiVOT with several state-of-the-art trackers.
        } 
        }
	\label{fig:LaSOT_AVisT_radar_plot}
  \vspace{-0.05in}
\end{figure*}

\begin{figure*}[!ht]
	\centering
 
        \begin{tabular}{ccc}

        \hspace{+0.5cm}
        
        {{\includegraphics[scale=0.40]{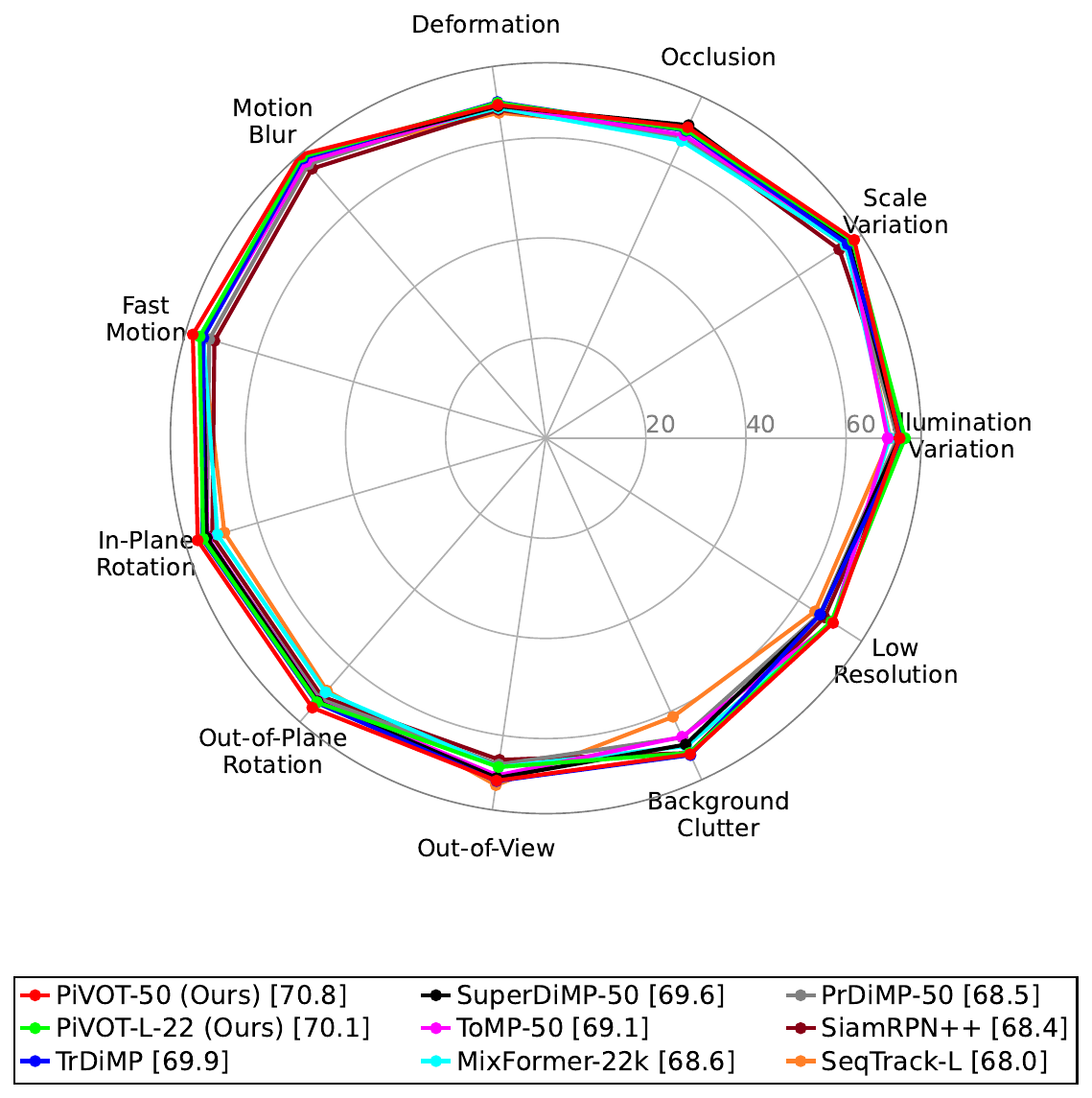}}}&
        
        
        {{\includegraphics[scale=0.40]{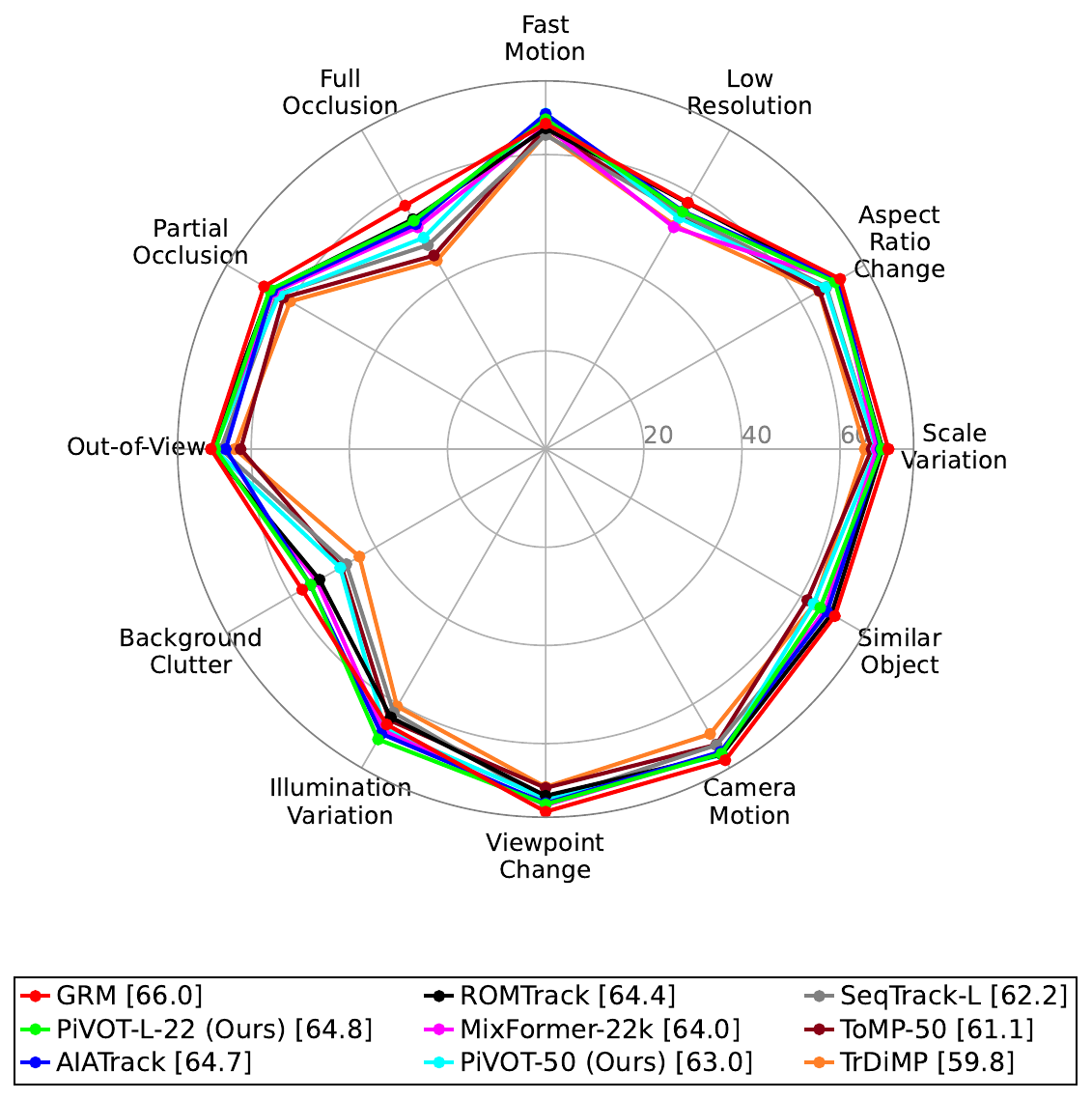}}}&

        \\

        (a) OTB-100 & 
        (b) UAV123 & 

        \end{tabular}
        
 
	\caption{
        \textbf{
        Attribute-based analysis of OTB-100 and UAV123, comparing PiVOT with several state-of-the-art trackers.
        } 
        }
	\label{fig:OTB_UAV_radar_plot}
\end{figure*}

%
\revone{
We analyze the attributes of the datasets through radar plots.
By conducting a detailed attribute analysis of these datasets using radar plots, we can enhance the understanding of our method relative to others and identify areas for improvement in future work.
Figure~\ref{fig:LaSOT_AVisT_radar_plot} and Figure~\ref{fig:OTB_UAV_radar_plot} provides details of this extensive evaluation, including 
many competing trackers~\cite{
GRM,
TATrack,
MixFormer,
SwinTrack,
AiATrack,
ROMTrack,
ToMP,
TrDiMP,
RTS,
STARK,
AlphaRefine,
TransT,
TrDiMP,
SiamRPN++,
SeqTrack}
and PiVOT.}

\revone{
In the analysis of LaSOT, as shown in Figure~\ref{fig:LaSOT_AVisT_radar_plot}(a), our PiVOT is more robust against Target Deformation and Fast Motion.
Regarding deformation, PiVOT brings the zero-shot category classification advantage of CLIP, making it more resilient to deformation.
The Fast Motion attribute indicates that the motion of the target object is larger than the size of its bounding box. A tracker that is more robust to this attribute typically demonstrates a better understanding of the scene, preventing reliance on the assumption that a target in a video usually moves slowly.
Additionally, our tracker also exhibits greater robustness to Viewpoint Change, Scale Variation, Partial Occlusion, etc. However, even though our tracker performs better in Full Occlusion, Fast Motion, Out-of-View, and Low Resolution compared to other trackers, there is still room for improvement.}

\revone{
In the analysis of AVisT, 
as depicted in Figure~\ref{fig:LaSOT_AVisT_radar_plot}(b), 
our PiVOT demonstrates greater robustness against 
Imaging Effects (images with noise), 
Camouflage (targets with similar appearance), 
Obstruction Effects (occlusion), 
and Weather Conditions (similar to Partial Occlusion) 
compared to other trackers. 
Although our PiVOT performs best in 
Target Effects
there remains room for improvement.
%
The Target Effects attribute assesses aspects like distractors, deforming objects, fast motion, and small targets. 
Although our PiVOT effectively handles distractors and deforming objects, 
it is limited in dealing with small targets due to the lack of sufficient semantic information that could be leveraged.
}

\revone{
The challenge of handling small-sized targets is also a common gap among most trackers in datasets like OTB-100 and UAV123, particularly in the Low-Resolution attribute, as shown in Figure~\ref{fig:OTB_UAV_radar_plot}(a) and Figure~\ref{fig:OTB_UAV_radar_plot}(b) respectively.
Moreover, similar to other trackers, there remains scope for improvement in how trackers manage occlusions.
}


\vspace{0.3in}
\section{Conclusion}
\label{sec:conclusion}

We introduce PiVOT, a promptable generic visual object tracker that leverages knowledge from foundation models, including CLIP~\cite{CLIP} and DINOv2~\cite{DINOv2}. The proposed prompt initialization mechanism, Prompt Generation Network, and Relation Modeling modules enable the tracker to incorporate visual prompts from the foundation model. PiVOT exploits CLIP for zero-shot knowledge transfer, where visual prompts are automatically generated by the aforementioned modules and further refined online by CLIP to guide the tracker toward the target of interest.
We further extend PiVOT by adopting the frozen ViT backbone from DINOv2 for feature extraction, thereby reducing inductive bias and computational cost while improving performance without fine-tuning the large ViT backbone on tracking data.
Comprehensive experiments and analyses on several challenging benchmarks demonstrate that PiVOT consistently improves tracking performance.

\bibliographystyle{IEEEtran}
\bibliography{main}
%
%
%
%
%
%
\newpage

\begin{IEEEbiography}[{\includegraphics[width=1in,height=1.25in,clip,keepaspectratio]{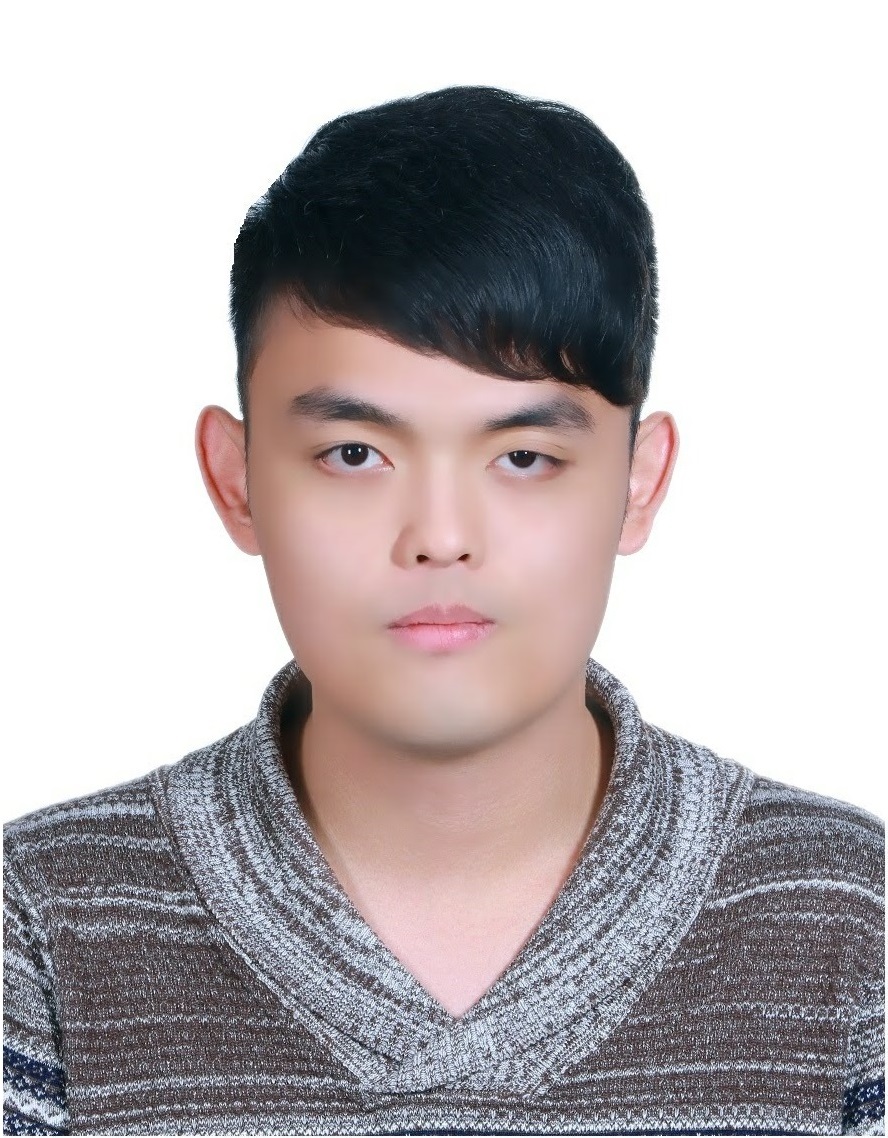}}]
{Shih-Fang Chen} 
is a PhD candidate in the Department of Computer Science at National Yang Ming Chiao Tung University, Taiwan. He received an MS degree in Computer Science and Engineering from Yuan Ze University, Taiwan, in 2020 and a BS degree from the Department of Computer Science and Information Engineering at Chaoyang University of Technology, Taiwan, in 2017. Since June 2020, he has been an honorary member of the Phi Tau Phi Scholastic Honor Society. His research primarily focuses on computer vision and deep learning.
\end{IEEEbiography}

\vspace{-10mm}

\begin{IEEEbiography}
[{\includegraphics[width=1in,height=1.25in,clip,keepaspectratio] {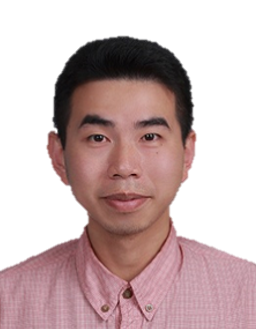}}]
{Jun-Cheng Chen} (Member, IEEE) is an Associate Research Fellow at the Research Center for Information Technology Innovation (CITI), Academia Sinica. He joined CITI as an assistant research fellow in 2019. He received the BS and MS degrees advised by Prof. Ja-Ling Wu in Computer Science and Information Engineering from National Taiwan University, Taiwan (R.O.C), in 2004 and 2006, respectively, where he received the PhD degree advised by Prof. Rama Chellappa in Computer Science from University of Maryland, College Park, USA, in 2016. From 2017 to 2019, he was a postdoctoral research fellow at the University of Maryland Institute for Advanced Computer Studies. His research interests include computer vision, machine learning, deep learning and their applications to biometrics, such as face recognition/facial analytics, activity recognition/detection in the visual surveillance domain, etc. His works have been recognized in prestigious journals and conferences in the field, including PNAS, TBIOM, CVPR, ICCV, ECCV, FG, WACV, etc. He was a recipient of the ACM Multimedia Best Technical Full Paper Award in 2006 and APSIPA ASC Best Paper Award in 2023.
\end{IEEEbiography}

\vspace{-10mm}

\begin{IEEEbiography}
[{\includegraphics[width=1in,height=1.25in,clip,keepaspectratio] {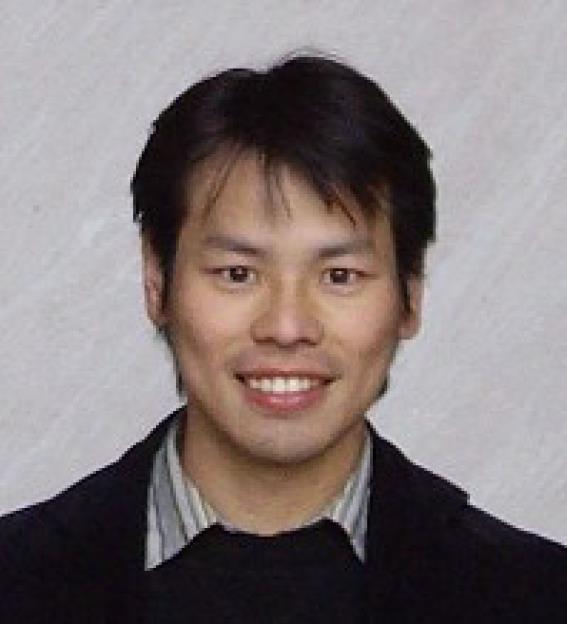}}]
{I-Hong Jhuo}  (Member, IEEE) is a senior applied scientist at Microsoft. 
He is an Active Participant in the Development of Innovative Technologies for information retrieval and recommendation systems while contributing to advances in the fields of computer vision, structured data and deep learning. His research interests include computer vision, information retrieval, and artificial intelligence. Recognized by awards: ACM Multimedia Grand Challenge 2012 and conducting the design of a top-performing video analytic system with NIST TRECVIDMED, Columbia University, New York, NY, USA.
\end{IEEEbiography}

\vspace{-10mm}

\begin{IEEEbiography}[{\includegraphics[width=1in,height=1.25in,clip,keepaspectratio]{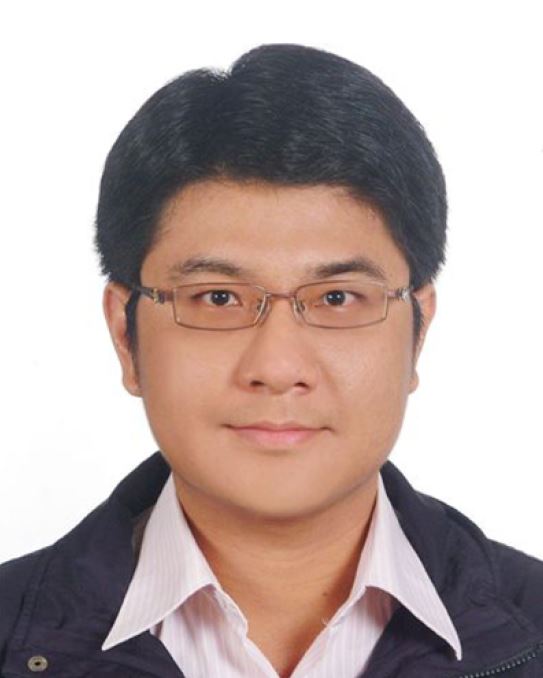}}]%
{Yen-Yu Lin} (Senior Member, IEEE) 
received the B.B.A. degree in Information Management, and the M.S. and Ph.D. degrees in Computer Science and Information Engineering from National Taiwan University, Taipei, Taiwan, in 2001, 2003, and 2010, respectively. He is currently a Distinguished Professor with the Department of Computer Science, National Yang Ming Chiao Tung University, Hsinchu, Taiwan. His research interests include computer vision, machine learning, and artificial intelligence. He serves as an Associate Editor of the International Journal of Computer Vision, Computer Vision and Image Understanding, and ACM Computing Surveys.
\end{IEEEbiography}

\end{document}